\documentclass[sigconf]{acmart}

\usepackage{amsmath,amsfonts,bm}

\def\eqref#1{(\ref{#1})}

\def\1{\bm{1}}

\newcommand{\Dr}{\mathcal{D_{\mathrm{r}}}}
\newcommand{\Df}{\mathcal{D_{\mathrm{f}}}}

\DeclareMathAlphabet{\mathsfit}{\encodingdefault}{\sfdefault}{m}{sl}
\SetMathAlphabet{\mathsfit}{bold}{\encodingdefault}{\sfdefault}{bx}{n}

\DeclareMathOperator*{\minimize}{\text{minimize}}

\newcommand{\btheta}{{\boldsymbol{\theta}}}

\newcommand{\bx}{\mathbf{x}}

\usepackage{pifont}

\usepackage{color, colortbl}
\definecolor{Gray}{gray}{0.93}
\definecolor{Orange}{rgb}{1,0.5,0}
\definecolor{DGray}{gray}{0.83}
\definecolor{LightCyan}{rgb}{0.88,1,1}

\usepackage[T1]{fontenc}

\usepackage{microtype}
\usepackage{graphicx}
\usepackage{subfigure}
\usepackage{booktabs}
\usepackage{hyperref}

\usepackage{wrapfig}
\usepackage{pifont}
\usepackage{color, colortbl}

\usepackage{blindtext}
\usepackage{lipsum}

\usepackage{multirow}
\usepackage{graphicx}
\usepackage{listings}

\usepackage{bbm}

\usepackage [english]{babel}
\usepackage [autostyle, english = american]{csquotes}

\usepackage{amsmath}

\usepackage{amssymb}
\usepackage{mathtools}
\usepackage{amsthm}

\usepackage[capitalize,noabbrev]{cleveref}

\theoremstyle{plain}

\theoremstyle{definition}

\theoremstyle{remark}

\usepackage[textsize=tiny]{todonotes}

\usepackage{mdframed}
\usepackage{pifont}
\usepackage{url}
\usepackage[most]{tcolorbox}
\usepackage{lipsum}
\usepackage{wrapfig}
\usepackage{booktabs}
\usepackage{multirow,mathtools } 

\usepackage{balance}
\usepackage{placeins}

\MakeOuterQuote{"}

\usepackage{hyperref}
\usepackage{url}

\everydisplay{\small}
\usepackage{siunitx}

\newcommand{\SL}[1]{\textcolor{purple}{SL: #1}}

\definecolor{Red}{RGB}{255,0,0}
\definecolor{Green}{RGB}{0,153,0}
\usepackage{float}

\AtBeginDocument{%
  }

\copyrightyear{2025}
\acmYear{2025}
\setcopyright{cc}
\setcctype{by-nd}
\acmConference[AISec '25]{Proceedings of the 2025 Workshop on Artificial Intelligence and Security}{October 13--17, 2025}{Taipei, Taiwan}
\acmBooktitle{Proceedings of the 2025 Workshop on Artificial Intelligence and Security (AISec '25), October 13--17, 2025, Taipei, Taiwan}\acmDOI{10.1145/3733799.3762973}
\acmISBN{979-8-4007-1895-3/2025/10}

\begin{document}

\title{LLM Unlearning on Noisy Forget Sets:\\
A Study of Incomplete, Rewritten, and Watermarked Data}



\author{Changsheng Wang}
\orcid{0009-0007-0957-638X}
\affiliation{%
  \institution{Michigan State University}
  \city{East Lansing}
  \country{USA}
}
\email{wangc168@msu.edu}

\author{Yihua Zhang}
\orcid{0000-0002-5147-9838}
\affiliation{%
  \institution{Michigan State University}
  \city{East Lansing}
  \country{USA}
}
\email{zhan1908@msu.edu}

\author{Dennis Wei}
\orcid{0000-0002-6510-1537}
\affiliation{%
  \institution{IBM Research}
  \city{San Jose}
  \country{USA}
}
\email{dwei@us.ibm.com}

\author{Jinghan Jia}
\orcid{0009-0001-7753-2326}
\affiliation{%
  \institution{Michigan State University}
  \city{East Lansing}
  \country{USA}
}
\email{jiajingh@msu.edu}

\author{Pin-Yu Chen}
\orcid{0000-0003-1039-8369}
\affiliation{%
  \institution{IBM Research}
  \city{San Jose}
  \country{USA}
}
\email{pinyuchen.tw@gmail.com}

\author{Sijia Liu}
\orcid{0000-0003-2817-6991}
\affiliation{%
  \institution{Michigan State University}
  \city{East Lansing}
  \country{USA}
}
\email{liusiji5@msu.edu}








\begin{abstract}
Large language models (LLMs) exhibit remarkable generative capabilities but raise ethical and security concerns by memorizing sensitive data, reinforcing biases, and producing harmful content. These risks have spurred interest in LLM unlearning, the task of removing knowledge associated with undesirable data from pre-trained models. However, most existing methods assume access to clean, well-defined forget data samples, whereas real-world forget data could often be low-quality, synthetically rewritten, or watermarked, casting doubt on the reliability of unlearning.
This work presents the first study of unlearning under perturbed or low-fidelity forget data, referred to as noisy forget sets. By systematically benchmarking state-of-the-art LLM unlearning methods, RMU and NPO, on such noisy forget sets, we find that unlearning remains surprisingly robust to perturbations, provided that core semantic signals are preserved. To explain this robustness, we propose a saliency-based interpretation: key semantic components that drive forgetting remain consistently influential despite substantial variation in surface form. This suggests that unlearning algorithms are primarily guided by deep semantic cues rather than shallow lexical patterns.
\end{abstract}




\maketitle

\section{Introduction}
\label{sec:intro}

Generative AI has been revolutionized by the advent of large language models (LLMs) \citep{touvron2023llama, achiam2023gpt, liu2024deepseek}. While their remarkable capabilities stem from training on massive and diverse datasets, LLMs also raise pressing ethical and security concerns. These include the potential leakage of private information through memorization \citep{sun2024trustllm,shi2024muse, chen2025safety}, the reinforcement of societal biases \citep{motoki2023more}, and the generation of harmful or illicit content \citep{wen2023unveiling,li2024wmdp}. Such challenges underscore the urgent need for effective methods to remove the influence of undesirable data from pre-trained models without compromising their utility, a task referred to as machine unlearning for LLMs, or simply \textbf{LLM unlearning} \citep{liu2024rethinking,yao2023large}.

Existing LLM unlearning methods typically assume access to a high-quality and well-defined forget dataset during training to obtain an unlearned model \cite{liu2024rethinking,li2024wmdp,maini2024tofu,shi2024muse}. However, this assumption often breaks down in real-world deployment scenarios, where the data targeted for removal is frequently noisy, incomplete, or synthetically generated~\cite{patel2024datadreamer, tang2023does, lupidi2024source2synth,liu2024query}. Therefore, a practical yet under-explored setting involves scenarios where sensitive content (such as copyrighted material) is either rewritten by LLMs into forgettable data formats or synthesized from coarse-grained unlearning concepts or knowledge.
We refer to the training forget sets that undergo such real-world \textit{perturbations}, including data incompleteness, rewriting, or watermarking, ``\textbf{noisy forget sets}''.
It is worth noting that our focus lies in natural perturbations of forget data during the training phase of unlearning, rather than in worst-case data poisoning scenarios \cite{goldblum2022dataset}.

Although recent efforts have examined the robustness of unlearning against {test-time} distribution shifts \citep{sun2024forget} or worst-case adversarial 
perturbations \citep{lynch2024eight,lucki2024adversarial,zhang2024generate}, to the best of our knowledge, no existing work has investigated the impact of \textit{train-time} noisy forget sets on the effectiveness of LLM unlearning. These noisy forget samples may introduce unintended artifacts, such as stylized phrasing or watermarking signals, that encode model-specific information~\citep{sun2024r,shu2024rewritelm}, potentially interfering with the unlearning process. Therefore, it is both important and timely to investigate this problem, as it directly impacts the applicability of current LLM unlearning methods to more realistic, low-quality forget sets.

Motivated by the above, the research question of this work is: \textbf{(Q) How do  noisy forget sets affect the effectiveness of LLM unlearning, even when evaluated on noiseless forget data?}

\begin{figure*}[htb] 
\vspace*{3mm}
\centering  
\begin{tabular}{cc}
 \includegraphics[width=0.45\textwidth]{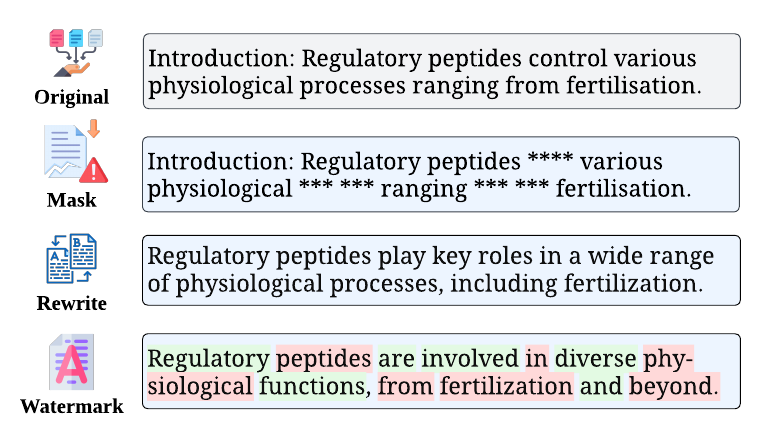} &
 \includegraphics[width=0.35\textwidth]{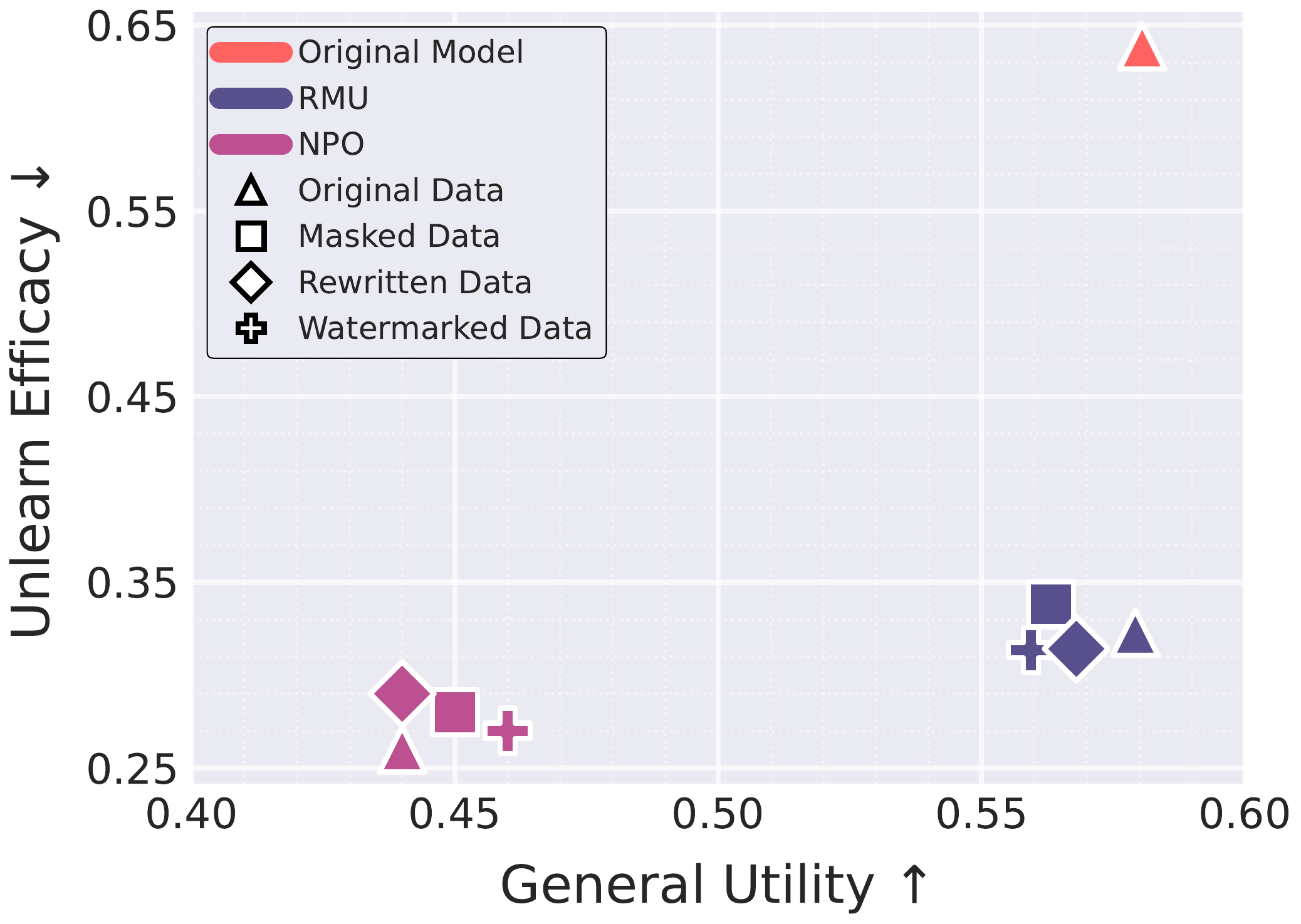}
\end{tabular}
\vspace*{-2mm}
\caption{\footnotesize{Illustrative examples of noisy forget data used during LLM unlearning training (left), and the performance (unlearning efficacy and utility) of the unlearned model evaluated on clean test data (right).
(Left) Different perturbation types applied to forget data during unlearning training. These include: \textbf{Mask}, where partial or missing content is simulated (masked tokens are indicated by \textbf{*}); \textbf{Rewrite}, where LLMs are prompted to generate semantically equivalent variants; and \textbf{Watermark}, where identifiable signals are embedded while preserving semantic meaning (tokens containing watermark signals are highlighted in red). \textbf{(Right)} Performance evaluation of two representative unlearning methods, NPO \cite{zhang2024negative} and RMU \cite{li2024wmdp}, applied to the Zephyr-7b-beta model on the WMDP dataset~\citep{li2024wmdp}. The forget data used for unlearning contains different types of perturbations (Mask, Rewrite, Watermark). \textit{Unlearn Efficacy} is reflected by the WMDP evaluation accuracy, where lower values indicate better unlearning performance. \textit{General Utility} reflects MMLU accuracy, where higher values indicate better retention of general model utility. Compared with unlearning on the original forget data format, different perturbation types have minimal impact on unlearning performance.}}
\label{fig:intro_overview}
\end{figure*}
To address (Q), we present the first systematic study of how the quality and structure of forget data influence LLM unlearning. Unlike prior works that often assume the availability of clean, curated forget sets, our investigation explicitly targets the more practical scenario in which forget data is perturbed through natural processes such as masking, rewriting, or watermarking. By conducting extensive experiments across multiple datasets and models, we demonstrate that unlearning performance remains robust to these perturbations, highlighting a surprising resilience of existing unlearning algorithms. \textbf{Fig.\,\ref{fig:intro_overview}} illustrates representative noisy forget set scenarios of the WMDP dataset \cite{li2024wmdp} and summarizes the resulting LLM unlearning performance. As shown by {Fig.\,\ref{fig:intro_overview}(right), unlearning with various types of perturbed forget data yields performance comparable to that achieved using the original clean forget set, both in terms of unlearning efficacy and general utility. Notably, all unlearned models exhibit substantially improved unlearning efficacy (reflected by lower accuracy on sensitive content) relative to the original model prior to unlearning.
To explain this robustness, we will propose a saliency-based interpretation: the core semantic components that drive forgetting are often preserved across perturbations, even when surface forms vary substantially. 

We summarize \textbf{our contributions} below:

    $\bullet$ We introduce a data-centric perspective to systematically analyze how noisy forget data, particularly LLM-generated or watermarked content, influences the unlearning process.

   $\bullet$ Using state-of-the-art LLM unlearning algorithms, including negative preference optimization (NPO) \cite{zhang2024negative} and representation misdirection unlearning (RMU) \cite{li2024wmdp}, we demonstrate through extensive experiments that the resulting unlearned models consistently achieve comparable performance, regardless of whether the forget data is watermarked, rewritten, or masked. See {Fig.\,\ref{fig:intro_overview}} for representative performance highlights.  
       
       $\bullet$ Through empirical and saliency-based analyses, it is shown that surface-level perturbations often high-saliency semantic elements, resulting in negligible degradation of unlearning effectiveness.

\section{Related Work}
\label{sec: related_work}
\subsection{Machine Unlearning in LLMs}  
Recent advances in machine unlearning for LLMs have shown promise in addressing risks associated with undesired data retention~\citep{liu2024rethinking,yao2024machine,zhuang2024uoe,maini2024tofu,eldan2023whos}. Practical implementations span critical applications, such as privacy protection through the removal of sensitive information~\citep{wu2023depn,yu2023unlearning}, prevention of harmful content generation~\citep{lu2022quark,li2024wmdp}, and elimination of memorized sequences~\citep{barbulescu2024each,jang2022knowledge}.
Most LLM unlearning methods rely on effective and efficient optimization techniques to avoid computationally prohibitive retraining while aiming to `faithfully' remove unwanted data-model influences~\citep{liu2024rethinking}. For instance, regularized optimization~\citep{yao2023large,liu2024rethinking,li2024wmdp,zhang2024negative} has been predominantly employed to balance unlearning effectiveness with preserved model utility post-unlearning. Some approaches employ localized interventions that target specific model components associated with unwanted capabilities~\citep{meng2022locating,wei2024assessing,jia2024wagle}.
Other unlearning approaches leverage in-context learning~\citep{pawelczyk2023context,thaker2024guardrail} or task vector~\citep{ilharco2022editing} to negate the effects of unwanted data or model capabilities in LLMs. While two recent studies~\citep{patil2025upcore, pal2025llm} have examined data-centric approaches to unlearning, their scope is limited to the coreset construction problem. In contrast, our work systematically investigates a wider spectrum of data perturbations.

\vspace*{-5mm}

\subsection{Robustness of Unlearning} 
While existing unlearning studies primarily focus on effectiveness under standard test conditions, recent work has begun to explore robustness under various \textit{test-time} perturbations. Some studies investigate test-time data corruptions, where models are evaluated on noisy or adversarially altered inputs~\citep{sun2024forget,schoepf2025redirection}. Others focus on adversarial robustness, such as jailbreak attacks where carefully crafted prompts can recover unlearned knowledge at test time~\citep{lucki2024adversarial, lynch2024eight, patil2023can}. Additionally, weight tampering robustness examines whether post-unlearning fine-tuning—on either small relevant or even irrelevant datasets—can undermine unlearning outcomes~\citep{deeb2024unlearning, hu2024jogging, lynch2024eight}. In contrast to these \textit{test-time} robustness perspectives, our work investigates a fundamentally different dimension: \textit{train-time} robustness. Specifically, we study how the quality of the forget set during training affects unlearning effectiveness. By introducing realistic, low-quality, and noisy forget sets, we examine the resilience of LLM unlearning methods against training-stage perturbations.
  

\subsection{Synthetic Data for LLMs}
Synthetic data refers to data that are artificially generated rather than directly collected from real-world events or human annotations. Early uses of synthetic data primarily relied on rule-based augmentation techniques, such as synonym replacement, random insertion or deletion, and back-translation~\citep{liu2024best, sennrich2015improving, wei2019eda}. However, rule-based augmentations often suffer from limited linguistic diversity and fail to introduce truly novel semantic patterns~\citep{yu2023large}. The advent of LLMs has significantly advanced synthetic data generation. LLMs, trained on massive corpora with billions of parameters, can produce high-quality, coherent, and human-like text when guided with carefully designed prompts~\citep{li2024data}. This prompt-based generation enables task-specific synthetic data creation with zero-shot settings~\citep{zubiaga2024natural}. Also, LLM watermarking represents a specialized form of synthetic data generation, where additional imperceptible information is embedded into the generated text during the synthesis process~\citep{kirchenbauer2023watermark, lee2023wrote, hu2023unbiased}. Unlike conventional synthetic data methods that focus solely on content diversity or task performance, watermark-based generation simultaneously serves both data augmentation and hidden signal embedding for IP protection. 


\section{Preliminaries and Problem Statement}
\label{sec: preliminary}


\subsection{A Primer on LLM Unlearning} 
Unlearning aims to remove the influence of undesired data from a trained model and its associated generation capabilities (such as producing sensitive or unsafe content), while preserving the model’s standard utility \citep{liu2024rethinking,eldan2023whos}. A typical unlearning setup involves a \textit{forget} objective that promotes the removal of specific information, and a utility-aware \textit{retain} objective that preserves overall model performance \citep{zhang2024negative,li2024wmdp,maini2024tofu}. The unlearning problem for LLMs can thus be formulated as:
\begin{align}
    \begin{array}{ll}
 \displaystyle \minimize_{\btheta}   & 
 \ell_{\mathrm{f}}(\btheta; \mathcal{D}_\mathrm{f}) + \gamma \ell_{\mathrm{r}}(\btheta; \mathcal{D}_\mathrm{r}),   
   \end{array}
   \label{eq: prob_LLM_unlearn}
\end{align}
where $\btheta$ denotes the model parameters to be optimized from a pre-trained state. The unlearning objective consists of a forget objective $\ell_{\mathrm{f}}$ and a retain objective $\ell_{\mathrm{r}}$. The parameter $\gamma \geq 0$ serves as a trade-off coefficient that balances forgetting and utility preservation.
When defining the forget and retain objectives in \eqref{eq: prob_LLM_unlearn}, it is typically assumed that one has access to a specific forget dataset $\mathcal{D}_\mathrm{f}$ and a retain dataset $\mathcal{D}_\mathrm{r}$. The forget dataset is often carefully curated, for example, fictitious author information in \cite{maini2024tofu}, book and news articles in \cite{eldan2023whos,shi2024muse}, and sensitive biosecurity-related content in \cite{li2024wmdp}.
In contrast, the retain dataset is usually selected with greater flexibility, such as using knowledge-related corpora or general-purpose utility datasets \cite{liu2024rethinking}.

Solving the LLM unlearning problem in \eqref{eq: prob_LLM_unlearn} follows a standard optimization framework, but its uniqueness primarily lies in the design of the forget objective $\ell_{\mathrm{f}}$. Two state-of-the-art LLM unlearning approaches are NPO (negative preference optimization) \cite{zhang2024negative} and RMU (representation misdirection unlearning) \cite{li2024wmdp}, where the former employs an untargeted, optimization divergence-driven approach, whereas the latter performs targeted unlearning by redirecting the representations of undesired data to random features.

Specifically,  NPO  \citep{zhang2024negative} instantiates the unlearning problem in \eqref{eq: prob_LLM_unlearn} with the  forget and retain objectives in \eqref{eq: NPO_forget_loss}, which reduces the model’s preference for the forget set $\Df$ by treating it analogously to negative responses in preference optimization \cite{rafailov2024direct}, but omitting the positive response. That is, 
\begin{align}
    \begin{array}{ll}
& \ell_{\mathrm{f}}(\btheta; \mathcal{D}_\mathrm{f}) =  \mathbb E_{(x,y) \in \Df} \left [    - \frac{2}{\beta} \log \sigma  \left ( - \beta \log \left ( \frac{\pi_{\btheta} (y | x) }{\pi_{\btheta_{\mathrm{o}}} (y | x)}\right ) \right ) 
    \right ] 
   \end{array}
   \label{eq: NPO_forget_loss}
\end{align}
where $\sigma (t) = 1/(1+e^{-t})$ is the sigmoid function, 
$\pi_{\btheta}$ represents the prediction probability of the model $\btheta$, and $\beta > 0 $ is a hyperparameter.
Minimizing the above forget loss drives the model to be unlearned $\pi_{\btheta}$ \textit{away} from the reference model $\pi_{\btheta_{\mathrm{o}}}$ (\textit{i.e.}, the original model $\btheta_{\mathrm{o}}$ before unlearning) given a forget sample in $\Df$. Here, $ x$ refers to the input and $y$ refers to the response. 
In NPO, the retain loss $\ell_{\mathrm{r}}$ is simply the prediction loss, \textit{i.e.}, the cross entropy loss between input $x$ and response $y$ within $\Dr$.

Additionally, {RMU}  \citep{li2024wmdp} enforces forgetting by mapping the hidden representations of the model $\btheta$ at a specific layer to random vectors on the forget set $\mathcal{D}_{\mathrm{f}}$, while simultaneously preserving
the original model’s representations $\btheta_{\mathrm{o}}$ on the retain set $\mathcal{D}_{\mathrm{r}}$. This leads to forget and retain objectives in \eqref{eq: prob_LLM_unlearn} as follows:
\begin{align}
\begin{array}{ll}
     \ell_{\text{f}} (\btheta; \mathcal{D}_{\mathrm{f}})
= \mathbb{E}_{\mathbf x \sim \mathcal{D}_{\mathrm{f}}} 
\left [  
\left\| M_{\btheta}(\mathbf x) - c \cdot \mathbf{u} \right\|_2^2 \right ]  \\
     \ell_{\mathrm{r}}  (\btheta; \mathcal{D}_{\mathrm{r}})
= \mathbb{E}_{\mathbf x \in \mathcal{D}_{\mathrm{r}}} 
\left[ 
\left\| M_{\btheta}(\mathbf x) - M_{\btheta_{\mathrm{o}}}(\mathbf x) \right\|_2^2 \right],   
\end{array}
\label{eq:rmu}
\end{align}
where 
$\| \cdot \|_2^2$ denotes the squared $\ell_2$ norm, 
$M_{\btheta}(\cdot)$ represents intermediate-layer representations of $\btheta$,  $\mathbf u$ is a random vector drawn from a standard
uniform distribution, and $c$ is a hyperparameter that controls the random vector scaling.


\subsection{Problem of Interest: LLM Unlearning on Noisy Forget Sets}
As motivated in Fig.\,\ref{fig:intro_overview}(left),  the forget set $\mathcal{D}_{\mathrm{f}}$ may be affected by various types of data ``noise''. Specifically, the forget set may contain: (1) masked samples resulting from missing or incomplete data; (2) rewritten examples generated by LLMs; and (3) watermarked content produced by   LLM watermarking at decoding for IP protection or ownership traceability.
To account for these scenarios, we propose an extended formulation of the unlearning problem, defined over a perturbed forget set $\mathcal{D}_\mathrm{f}^\prime$ that captures various real-world corruption cases:
\begin{align}
    \begin{array}{ll}
 \displaystyle \minimize_{\btheta}   & 
 \ell_{\mathrm{f}}(\btheta; \mathcal{D}_\mathrm{f}^\prime) + \gamma \ell_{\mathrm{r}}(\btheta; \mathcal{D}_\mathrm{r}).
   \end{array}
   \label{eq:perturbed_unlearn}
\end{align}
 Our goal is to investigate how a noisy forget set $\mathcal{D}_\mathrm{f}^\prime$ affects the unlearning performance of the model optimized by \eqref{eq:perturbed_unlearn}. Since our focus is on the sensitivity of unlearning to ``noise'' in the forget data, we consider perturbations only in the forget set, while keeping the retain set unchanged.
 In the next section, we detail the construction of noisy forget sets in our evaluation.

\section{Forget Data ``Noise'' in LLM Unlearning}

In this section, we introduce three practical scenarios that may give rise to noisy forget sets ($\mathcal{D}_\mathrm{f}^\prime$) in LLM unlearning. These scenarios reflect common data quality issues, including incomplete, rewritten, and watermarked forget data, which unlearning algorithms are likely to encounter in real-world settings.


\subsection{Masked Forget Data}
The first noisy scenario is \textit{mask} of forget data, where only partial information is available for unlearning. This setting commonly arises when the full sensitive content cannot be accessed or disclosed. For instance, in WMDP \citep{li2024wmdp}, which aims to remove biosecurity-related hazardous knowledge from LLMs, only fragments of each article are provided for unlearning. A similar situation occurs when forgetting copyrighted content from books \citep{shi2024muse,eldan2023whos}, where only selected paragraphs may be included, resulting in incomplete and partially masked forget data.

%

This scenario is also related to the coreset-based unlearning framework \cite{pal2025llm}, where only a small subset of the forget dataset is used to achieve lossless unlearning. However, the key distinction lies in the granularity of incompleteness: coreset methods select a subset of forget {samples}, whereas the incompleteness we consider here applies at the level of each individual forget {sample}, where only partial content from a full data instance is provided for unlearning.
To be concrete, we define a token-level masking function $\textsc{Mask}_{\delta}(\mathbf{x})$, which denotes that for each forget sample $\mathbf{x} \in \mathcal{D}_{\mathrm{f}}$, $\delta$ (\%) of its tokens are masked, with the masked positions selected uniformly at random.
This yields the following noisy forget set:
\begin{align}
\begin{array}{l}
\mathcal{D}^\prime_{\mathrm{f}} = \left\{ \textsc{Mask}_\delta(\bx) \mid \forall \bx \in \mathcal{D}_\mathrm{f} \right\}.   
   \end{array}
   \label{eq: incomplete_data}
\end{align}
We refer readers to Fig.\,\ref{fig:intro_overview}(left) for example of masked forget data.

\textbf{Fig.\,\ref{fig:mask_ratio}} shows that unlearning performance remains largely unaffected as the mask ratio increases, up to approximately 30\% of the forget data being masked. Beyond this threshold, a noticeable degradation in unlearning efficacy is observed. This suggests that 30\% is likely the highest level of masking that does not significantly compromise unlearning performance {on the WMDP dataset}. Unless otherwise specified, we adopt a 30\% mask ratio as the default setting for masked forget data.


\begin{figure}[htb]
    \centering
    \includegraphics[width=0.45\textwidth]{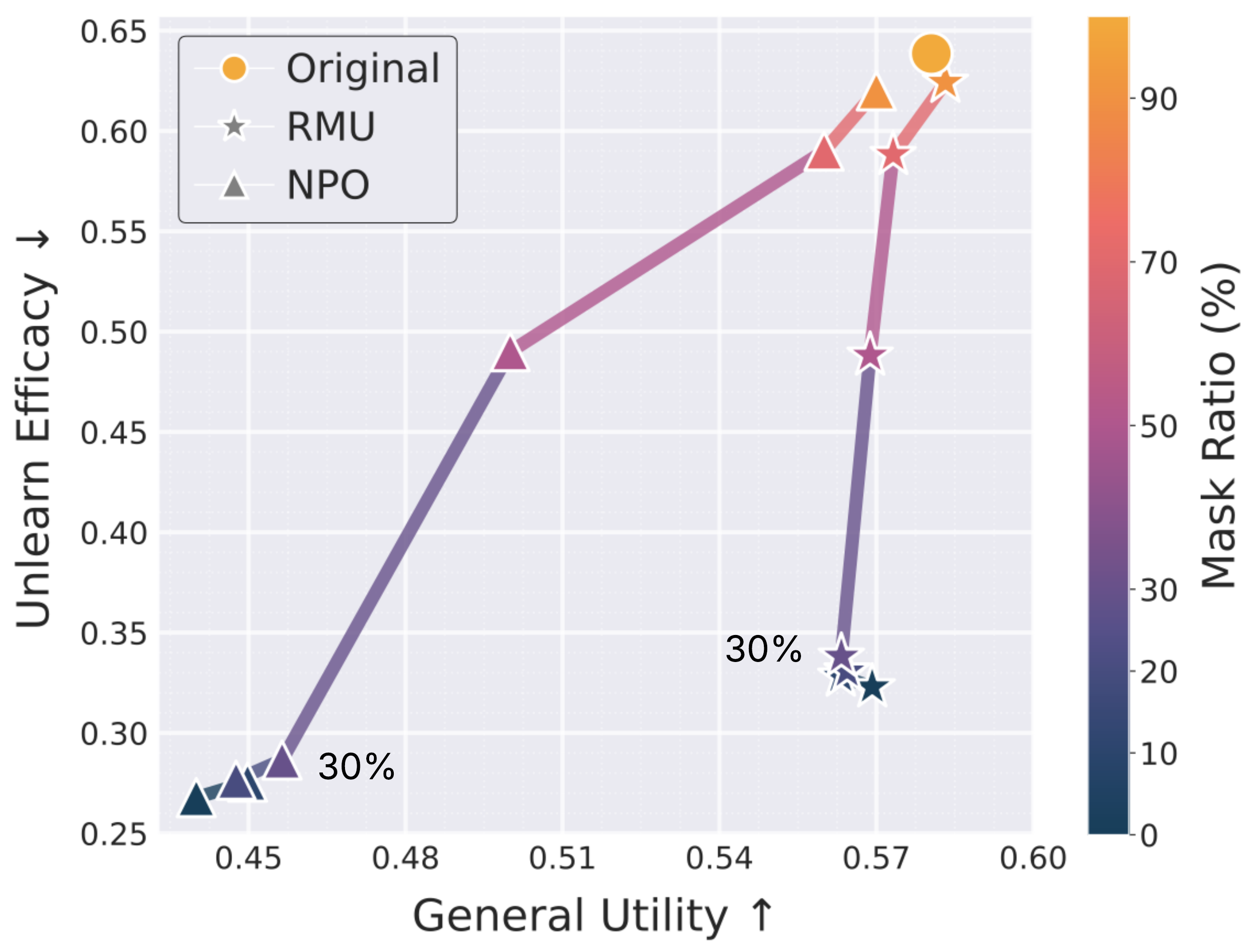}
    \vspace*{-1.5em}
    \caption{\small{Impact of masking ratio on unlearning performance across two representative unlearning methods, NPO and RMU, applied to the Zephyr-7b-beta model on the WMDP dataset~\citep{li2024wmdp}, where the masking ratio ($\delta$) varies from 0\% to 90\%. Here,  0\% corresponds to the original, unmasked forget data. The unlearning performance is measured by \textit{Unlearn Efficacy} and \textit{General Utility} as shown in Fig.\,\ref{fig:intro_overview}.}}
    \vspace*{-1.5em}
    \label{fig:mask_ratio}
\end{figure}

\subsection{Rewritten Forget Data}
The second noisy scenario arises when the forget data is rewritten by LLMs. This typically occurs in settings where unlearning is performed over synthetic forget data, either because the original information is only available in high-level conceptual form, or due to the need to regenerate underspecified content. In such cases, full forget examples are synthesized based on coarse descriptions, and unlearning is conducted using these generated data.

To simulate this scenario, we prompt the target LLM (the model before unlearning) to produce semantically preserved rewrites of each forget example. This allows us to assess how rewriting-induced variation affects unlearning efficacy.
Let $\textsc{Rewrite}(\cdot)$ denote a rewriting function that generates a paraphrased variant of a forget sample $\bx$ while preserving its original semantics.
This yields the noisy forget set
\begin{align}
\begin{array}{l}
\mathcal{D}_\mathrm{f}^\prime = \left\{\textsc{Rewrite}(\bx) \mid \forall \bx \in \mathcal{D}_\mathrm{f} \right\}.   
   \end{array}
   \label{eq: rewrite_data}
\end{align}
To construct the rewritten forget data in \eqref{eq: rewrite_data}, we ensure semantic consistency and preservation of the original intent. This process follows constraints similar to those in back-translation and controlled paraphrasing techniques commonly used in text generation. The exact prompt used for generating the rewrites is provided below. In brief, it instructs the language model to rewrite a given input text while maintaining its original meaning and enhancing clarity, coherence, and conciseness.
We refer readers to Fig.\,\ref{fig:intro_overview}(left) for the example of rewritten forget data. Here is the prompt: 


\vspace{2mm}
\begin{tcolorbox}[colback=gray!5, colframe=gray!50!black, title=Rewrite prompt]
\small{
\textbf{Prompt:} You are an AI language model tasked with rewriting the following text. Your goal is to maintain the original meaning while improving clarity, coherence, and conciseness. Ensure the rewritten text sounds natural and fluent. Do not add new information or change the intended message. \\
\textbf{Original Text:} \{original text\}
}
\label{main:prompt}
\end{tcolorbox}


\subsection{Watermarked Forget Data} 
Watermarked LLMs are models whose decoding processes are deliberately altered to embed imperceptible patterns into generated text, enabling ownership verification and content attribution~\citep{kirchenbauer2023watermark, dathathri2024scalable, wu2023dipmark, zhao2023provable}. These techniques are increasingly adopted by LLM providers to deter misuse in content-generation scenarios. However, they pose an unknown challenge for LLM unlearning: when synthetic or rewritten forget data is produced by a watermarked LLM, the resulting text contains watermark artifacts, rendering the forget set $\Df$ inherently `noisy'.

Thus, we consider the scenario where the forget data is rewritten using a watermarked LLM, following the rewriting procedure used earlier. This process introduces watermark signals into the rewritten text due to the modified decoding behavior of the watermarked model. Consequently, the resulting forget set contains both semantic transformations and imperceptible watermark patterns. Formally, we denote this watermarked forget set as:
\begin{align}
\begin{array}{l}
\mathcal{D}_\mathrm{f}^\prime = \left\{\textsc{Watermark}(\bx) \mid \forall \bx \in \mathcal{D}_\mathrm{f} \right\}  ,
   \end{array}
   \label{eq: watermark_data}
\end{align}
where $\textsc{Watermark}(\bx)$ denotes the output of a watermark-enabled LLM decoding process for input $\bx$.

Following prior LLM watermarking literature~\citep{kirchenbauer2023watermark, dathathri2024scalable}, we consider two representative watermarking strategies:
(1) \textbf{Logits-based watermarking} (\textit{e.g.}, KGW~\citep{kirchenbauer2023watermark}), which modifies token selection probabilities during decoding to embed hidden patterns. Specifically, this approach perturbs the model’s logits before sampling. In KGW, the vocabulary is partitioned at each decoding step into a ``green list'' $G$ and a ``red list'' $R$, determined by a seeded hash of the previous token. A fixed bias $\delta$ is then added to the logits of tokens in the red list to subtly steer generation. The modified logit $\tilde{l}^{(t)}_k$ for token $k$ at step $t$ is computed as:
\begin{align}
\tilde{l}^{(t)}_k =
\begin{cases}
    l^{(t)}_k + \delta, & \text{if } k \in R \\
    l^{(t)}_k, & \text{if } k \in G,
\end{cases}
\label{eq: lk_eq}
\end{align}
where $l^{(t)}_k$ denotes the original logit. 
This logit modification introduces a biased generation, increasing the likelihood of selecting certain tokens from the red list. And
the hardness parameter $\delta > 0$ controls the strength of the watermark signal: larger $\delta$ values increase watermark detectability but may degrade generation quality. 

(2) \textbf{Sampling-based watermarking} (\textit{e.g.}, SynthID~\citep{dathathri2024scalable}) embeds watermark signals by modifying the token sampling procedure without altering model logits. A representative method is SynthID, which employs a multi-layer Tournament Sampling scheme. At each generation step $t$, a candidate pool $\mathcal{Y}_t$ of $N$ tokens is sampled (with possible repeats) from the model’s output distribution. An $m$-layer tournament then selects the output token $x_t$: in each layer $\ell \in {1, \ldots, m}$, token pairs are scored using a pseudorandom function $g_{\ell}(x, r_t)$, with winners advancing to the next round. Here, the scoring function is typically stochastic (\textit{e.g.}, Bernoulli or Uniform), and the randomness seed $r_t$ is deterministically derived from the context and a secret key. The watermark strength is governed by the number of tournament layers $m$: higher $m$ increases watermark detectability but reduces sampling entropy, potentially affecting output fluency.




\definecolor{Green}{RGB}{0,150,0}     
\definecolor{Blue}{RGB}{0,0,200}      
\definecolor{Red}{RGB}{220,0,0}

\definecolor{mygreen}{RGB}{200, 255, 200}  
\definecolor{myred}{RGB}{255, 200, 200}    
\definecolor{mypurple}{RGB}{230, 200, 255}

\setlength{\fboxsep}{0.5pt}

\newcommand{\redbox}[1]{\colorbox{myred!30}{\strut #1}}
\newcommand{\greenbox}[1]{\colorbox{mygreen!30}{\strut #1}}
\newcommand{\purplebox}[1]{\colorbox{mypurple!30}{\strut #1}}

\begin{table}[htp]
\centering
\caption{\small{
Examples of watermarked forget data under varying watermark strengths, where samples processed by two representative watermarking methods, KGW~\citep{kirchenbauer2023watermark} and SynthID~\citep{dathathri2024scalable}. For KGW, tokens highlighted in \redbox{red} belong to the \textbf{red list}, and those in \greenbox{green} belong to the \textbf{green list}; a higher proportion of red tokens reflects a stronger watermark signal. For SynthID, \purplebox{purple}-highlighted tokens mark positions selected through multi-layer tournament sampling; denser purple highlights indicate stronger watermarking and greater watermark detectability.
}}
\label{tab:wm_example}
\scalebox{0.9}{
\begin{tabular}{c|c|c}
\toprule[1pt]
\midrule
\multicolumn{2}{c|}{\shortstack{\textbf{KGW} \\ \textbf{($\delta=2$)}}}  & 
\parbox[c]{6cm}{\vspace{0.5ex}
\noindent
\greenbox{These} \redbox{peptides} \greenbox{often} \greenbox{act} \greenbox{as} \redbox{signaling} \redbox{molecules}\greenbox{,} \redbox{coord}\redbox{inating} \redbox{responses} \greenbox{to} \redbox{internal} \greenbox{and} \greenbox{extern}\redbox{al} \redbox{stimuli}\greenbox{.} \redbox{The} \redbox{study} \greenbox{of} \redbox{regulatory} \redbox{peptides} \redbox{involves} \redbox{understand}\redbox{ing} \greenbox{their} \greenbox{synthe}\redbox{sis}\greenbox{,} \greenbox{process}\redbox{ing}\greenbox{,} \greenbox{and} \redbox{function}\greenbox{,} \greenbox{as} \greenbox{well} \greenbox{as} \greenbox{their} \redbox{interactions} \greenbox{with} \redbox{receptors} \greenbox{.} \vspace{0.5ex}
} \\
\midrule
\multicolumn{2}{c|}{\shortstack{\textbf{KGW} \\ \textbf{($\delta=6$)}}}  & 
\parbox[c]{6cm}{\vspace{0.5ex}
\noindent
\redbox{Peptids} \greenbox{is} \greenbox{like} \redbox{very} \redbox{im}\redbox{portant} \greenbox{for} \redbox{signal} \redbox{thing}. \greenbox{They} \redbox{making} \redbox{some}\redbox{thing} \redbox{inside} \greenbox{and} \redbox{outside}. \redbox{Also} \greenbox{they} \greenbox{do} \redbox{many} \redbox{\textbf{job}} \greenbox{like} \redbox{\textbf{job}} \redbox{of} \redbox{signal} \greenbox{and} \redbox{another} \redbox{\textbf{job}}. \redbox{People} \redbox{study} \greenbox{for} \redbox{why} \redbox{peptides} \redbox{happen} \greenbox{and} \redbox{make} \greenbox{and} \redbox{also} \redbox{how} \redbox{make} \greenbox{and} \redbox{then} \redbox{doing} \redbox{thing} \greenbox{with} \redbox{receptor} \redbox{or} \redbox{other} \redbox{stuff}. \redbox{Function} \greenbox{is} \redbox{also} \redbox{some}\redbox{thing} \redbox{about}.\vspace{0.5ex}
} \\
\midrule
\multicolumn{2}{c|}{\shortstack{\textbf{SynthID} \\ \textbf{($m=2$)}}}  & 
\parbox[c]{6cm}{\vspace{0.5ex}
\noindent
\purplebox{Regulat}\purplebox{ory} peptides play \purplebox{cruci}\purplebox{al} roles as \purplebox{signal}\purplebox{ing} \purplebox{molecule}\purplebox{s}, \purplebox{mediat}\purplebox{ing} various \purplebox{physio}\purplebox{logical} \purplebox{responses}\purplebox{s} to both internal and external stimuli. Research on these peptides \purplebox{focus}\purplebox{es} on \purplebox{understand}\purplebox{ing} their \purplebox{bio}\purplebox{synthesis}. \vspace{0.5ex}
} \\ 
\midrule
\multicolumn{2}{c|}{\shortstack{\textbf{SynthID} \\ \textbf{($m=6$)}}}  & 
\parbox[c]{6cm}{\vspace{0.5ex}
\noindent
\purplebox{Regulatory} are \purplebox{important} \purplebox{because} \purplebox{they} \purplebox{help} \purplebox{with} \purplebox{signals}. \purplebox{They} \purplebox{do} \purplebox{things} \purplebox{inside} \purplebox{the} \purplebox{body} and \purplebox{outside} too. \purplebox{Studying} \purplebox{peptides} is \purplebox{about} \purplebox{knowing} \purplebox{how} \purplebox{they} \purplebox{are} \purplebox{made}, \purplebox{how} \purplebox{they} \purplebox{work}, and \purplebox{how} \purplebox{they} \purplebox{connect} \purplebox{with} \purplebox{receptors} and \purplebox{other} \purplebox{molecules}. \vspace{0.5ex}
} \\
\midrule
\bottomrule[1pt]
\end{tabular}
}
\end{table}

As shown in \textbf{Table\,\ref{tab:wm_example}}, KGW-based watermarking is reflected by the proportion of red-list tokens within the generated text. As the watermark strength increases (from $\delta=2$ to $\delta=6$), a greater fraction of tokens are forced into the red list, making the watermarking stronger (and thus more detectable). However, at $\delta=6$, the strong bias towards red-list tokens noticeably degrades text quality, as reflected by repetitive and unnatural token usage (\textit{e.g.}, repeated occurrences of ``job'' in the KGW $\delta=6$ example). Unlike KGW, which modifies token logits to manipulate sampling probabilities, SynthID embeds watermarks by altering the sampling process through multi-layer tournament selection guided by a hidden key. Increasing the number of tournament layers from $m=2$ to $m=6$  improves watermark strength, but with less impact on fluency and coherence compared to KGW.

\section{Experiments}
\label{sec: exp}

\subsection{Experiment Setups}

\subsubsection{LLM unlearning datasets, models, and methods}

We conduct experiments on two representative LLM unlearning benchmarks: \textbf{WMDP}~\citep{li2024wmdp} and \textbf{MUSE}~\citep{shi2024muse}. The \textbf{WMDP} dataset focuses on removing hazardous domain knowledge from the biosecurity domain, evaluated on the \texttt{Zephyr-7B-beta} model~\citep{tunstall2023zephyr}. The \textbf{MUSE} benchmark includes the task of removing memorized content from the Harry Potter book series (labeled as "Books"). For Books, we use the \texttt{ICLM-7B} model from \citep{shi2024muse}, which follows the setting in MUSE benchmark. 


For unlearning methods, we select two state-of-the-art baselines: \textbf{NPO}~\citep{zhang2024negative} and \textbf{RMU}~\citep{li2024wmdp}, both formulated under the general LLM unlearning objective defined in Eq.\,\eqref{eq: prob_LLM_unlearn}. On the \textbf{WMDP} dataset, we exclusively run \textbf{RMU}, as it represents the state-of-the-art unlearning method specifically designed for hazardous knowledge removal in this setting. On the \textbf{MUSE} benchmark, we adopt \textbf{NPO} due to its leading performance reported in the MUSE benchmark. This selective choice aligns with prior work~\citep{shi2024muse} 
and ensures fair  comparisons under consistent evaluation settings.


\subsubsection{Evaluation metrics}

We evaluate unlearning performance from two complementary perspectives: \textit{Unlearn Efficacy \textbf{(UE)}}, which measures the extent to which target knowledge has been removed, and \textit{Utility \textbf{(UT)}}, which assesses the retention of general, unlearning-irrelevant knowledge. For \textbf{WMDP}~\citep{li2024wmdp}, UE is measured by the \textit{question-answering accuracy} on the designated biosecurity forget set. A lower question-answering accuracy after unlearning indicates better unlearning efficacy. UT on WMDP is evaluated using the zero-shot performance on the MMLU benchmark~\citep{hendrycks2020measuring}, which reflects the preservation of the model’s general knowledge and standard generation abilities unrelated to the forget set. For \textbf{MUSE}~\citep{shi2024muse}, we follow the benchmark’s recommended multi-metric evaluation protocol for UE, including three distinct metrics: (1) \textit{Verbatim Memorization \textbf{(VerbMem)}}, which measures the model's next-token prediction accuracy on the forget set $\Df$, indicating its ability to reproduce memorized sequences; (2) \textit{Knowledge Memorization \textbf{(KnowMem)}}, which evaluates the model’s ability to answer knowledge-based questions derived from the undesired forget set content; and (3) \textit{Privacy Leakage \textbf{(PrivLeak)}}, which quantifies the risk of membership inference, measuring the model’s tendency to reveal whether specific data points from $\Df$ were present in the original training set. Lower VerbMem and KnowMem scores indicate better unlearning efficacy, while PrivLeak values closer to zero reflect reduced membership leakage. UT on MUSE is evaluated by reporting \textit{KnowMem} on the benchmark’s retain set $\Dr$, where higher KnowMem indicates better utility preservation on non-target knowledge. 

Additionally, to analyze residual memorization patterns and the consistency of forgetting behavior across different unlearning runs or perturbations of the forget data, we report \textit{Error Set Overlap \textbf{(ESO)}}~\citep{li2024wmdp}, which quantifies the semantic alignment of forget set error patterns between different unlearned models. Higher ESO indicates more consistent forgetting across methods or perturbation settings. For different types of data perturbations, we use the following notations for clarity throughout the experiments. \textbf{Mask} denotes the masked forget set, with a default masking ratio of 30\%. \textbf{Rewrite} refers to the rewritten forget set generated via semantic rewriting prompts. \textbf{WM(KGW)} indicates the logits-based watermarking method KGW with $\delta$=2, and \textbf{WM(SynthID)} refers to the sampling-based SynthID method with m=4.  It is worth noting that the chosen watermarking strengths reflect a favorable trade-off between watermark detectability and the preservation of text quality.



\subsection{Experiment Results}

\subsubsection{LLM unlearning on masked forget data} 
Recall from Fig.\,\ref{fig:mask_ratio} that we analyze how varying the mask ratio affects unlearning performance for both RMU and NPO. When the mask ratio increases from 0\% to 30\%, unlearning efficacy remains largely stable, suggesting that moderate input masking does not significantly impede the forgetting of target knowledge. However, beyond the 30\% threshold, a clear degradation in unlearning performance emerges, likely due to over-masking of forget data content in the forget set, which limits the model’s ability to identify and erase relevant knowledge. This is supported by the observation that higher mask ratios are also associated with a slight improvement in general utility, 
potentially because less specific knowledge is removed, thereby reducing collateral forgetting and preserving overall model performance.

\subsubsection{LLM unlearning versus watermarking strength.}

\textbf{Table\,\ref{tab:watermark_strength_results}}  presents
how increasing watermarking strength impacts LLM unlearning performance. For the logits-based KGW method, performance begins to degrade noticeably at $\delta=4$, with both unlearning efficacy and general utility metrics worsening relative to $\delta=2$ and the clean-data baseline. As shown in Table\,\ref{tab:wm_example}, higher $\delta$ values introduce stronger token-level biases, which reduce text quality and make the forget set less informative for guiding effective unlearning. This is reflected in worse UE scores at $\delta=6$ vs. $\delta=2$ or  $m=6$ vs. $m=2$.
Compared to KGW, the sampling-based SynthID method maintains relatively more stable performance as $m$ increases. However, at $m=6$, the deeper tournament sampling introduces more severe text distortion, leading to worse unlearning efficacy. These results suggest that while stronger watermarking increases watermark detectability, it may compromise unlearning effectiveness, particularly at relatively large watermark strengths.



\begin{table}[htb]
\centering
\caption{Unlearning performance under different watermarking strengths. This table reports the unlearning performance of two representative unlearning methods, RMU and NPO, applied to the Zephyr-7b-beta model on the WMDP dataset~\citep{li2024wmdp}. Both \textit{Unlearn Efficacy} (UE ↓) and \textit{General Utility} (UT ↑) are evaluated across forget sets perturbed by different watermarking strategies. Two watermarking methods are considered: logits-based watermarking (KGW) with varying $\delta$ values ($\delta$=2, 4, 6) and sampling-based watermarking (SynthID) with different tournament depths ($m$=2, 4, 6). 
}
\label{tab:watermark_strength_results}
\resizebox{0.44\textwidth}{!}{
\begin{tabular}{c|cc|cc}
\toprule
\multirow{2}{*}{\parbox[c]{2.5cm}{\centering \textbf{Watermark}\\\textbf{Strength}}} & \multicolumn{2}{c|}{\textbf{RMU}} & \multicolumn{2}{c}{\textbf{NPO}} \\
\cline{2-5}
& \textbf{UE~$\downarrow$} & \textbf{UT~$\uparrow$} & \textbf{UE~$\downarrow$} & \textbf{UT~$\uparrow$} \\
\midrule
Original Model & 0.6386 & 0.5805 & 0.6386 & 0.5805 \\
\midrule
Original Data     & 0.3229 & 0.5692 & 0.2603 & 0.4436 \\
\midrule
\multicolumn{5}{c}{\textbf{Logits-based Watermarking (KGW)}} \\
\midrule
$\delta=2$     & 0.3134 & 0.5694 & 0.2765 & 0.4521 \\
$\delta=4$     & 0.3652 & 0.5631 & 0.3124 & 0.4675 \\
$\delta=6$     & 0.3764 & 0.5461 & 0.3265 & 0.4613 \\
\midrule
\multicolumn{5}{c}{\textbf{Sampling-based Watermarking (SynthID)}} \\
\midrule
$m=2$          & 0.3201 & 0.5673 & 0.2675 & 0.4498 \\
$m=4$          & 0.3221 & 0.5684 & 0.2641 & 0.4501 \\
$m=6$          & 0.3465 & 0.5512 & 0.2945 & 0.4598 \\
\bottomrule
\end{tabular}
}
\vspace*{-1em}
\end{table}

\subsubsection{Unlearning performance under perturbed forget data.}  
\textbf{Table\,\ref{tab:perturbation_results}} and \textbf{Table\,\ref{tab:perturbation_results_muse}} summarize the unlearning performance of RMU and NPO on the WMDP and MUSE benchmarks, respectively. For RMU on WMDP, unlearning significantly improves efficacy over the original model (no unlearning) with only modest utility loss. Perturbations to the forget set, including incomplete masking, rewriting, and watermarking, result in only minor variations. Incomplete masking slightly reduces utility, likely due to the removal of key semantic tokens, while rewritten and watermarked forget sets achieve comparable unlearning and utility as the clean baseline. For NPO on MUSE, all variants achieve near-complete removal of verbatim memorization and strong suppression of privacy leakage, with minimal utility degradation. These results across methods and benchmarks suggest that unlearning is generally robust to noisy forget sets.

\begin{table}[htb]
\centering
\caption{Performance of RMU unlearning on perturbed forget data using Zephyr-7B-beta.
Comparison of unlearning efficacy and general utility on the WMDP benchmark under various forget data conditions: no unlearning (\textit{i.e.}, original model before unlearning on forget data), clean, randomly masked (incomplete), semantically rewritten (prompt-based), and watermarked (KGW and SynthID).
}
\label{tab:perturbation_results}
\vspace*{-3mm}
\resizebox{0.6\linewidth}{!}{
\begin{tabular}{c|c|c}  
\toprule[1pt]
\textbf{Forget data type} & \textbf{UE $\downarrow$} & \textbf{UT $\uparrow$} \\
\midrule
No unlearning     & 0.6386 & 0.5805 \\
\midrule
Clean data                & 0.3229 & 0.5692 \\
\midrule
Mask      & 0.3382 & 0.5632 \\
Rewrite         & 0.3142 & 0.5680 \\
WM (KGW)        & 0.3134 & 0.5694 \\
WM (SynID)  & 0.3221 & 0.5684 \\
\bottomrule[1pt]
\end{tabular}}
\end{table}


\begin{table}[htb]
\centering
\caption{\small{
Unlearning performance of NPO on MUSE-Books using ICLM-7B under various forget data perturbations.
It reports UE across Verbatim Memorization, Knowledge Memorization, and Privacy Leakage, along with  UT measured as retained performance on Knowledge Memorization. Forget set variants include clean forget data, randomly masked (incomplete), semantically rewritten (prompt-based), and watermarked versions (KGW and SynthID).
}}
\label{tab:perturbation_results_muse}
\vspace*{-3mm}
\resizebox{0.46\textwidth}{!}{
\begin{tabular}{c|ccc|c}
\toprule
\multirow{3}{*}{\textbf{Forget data type}} & \multicolumn{3}{c|}{\textbf{UE}} & \textbf{UT} \\
\cmidrule(lr){2-4} \cmidrule(lr){5-5}
& 
\begin{tabular}{c}
\textbf{VerbMem} \\
($\downarrow$)
\end{tabular} & 
\begin{tabular}{c}
\textbf{KnowMem} \\
($\downarrow$)
\end{tabular} & 
\begin{tabular}{c}
\textbf{PrivLeak} \\
($\rightarrow 0$)
\end{tabular} & 
\begin{tabular}{c}
\textbf{KnowMem} \\
($\uparrow$)
\end{tabular} \\
\midrule
No unlearning        & 99.80 & 59.40 & -57.50 & 66.90 \\
\midrule
Clean data   &  0.00 &  1.18 & -42.07 & 57.19 \\
\midrule
Mask             &  0.05 &  0.33 & -49.36 & 55.31 \\
Rewrite                &  0.06 &  0.00 & -53.43 & 50.73 \\
WM(KGW)                &  0.12 &  1.00 & -53.51 & 56.92 \\
WM(SynthID)            &  0.05 &  1.13 & -48.65 & 56.42 \\
\bottomrule
\end{tabular}
}
\end{table}

\subsubsection{Analyzing error set overlap to assess unlearning robustness.} 
To further validate that data perturbations do not compromise the core forgetting objective, we examine whether unlearned models erase the same underlying knowledge across forget set variants. Specifically, we compare the sets of incorrectly answered WMDP questions for models unlearned with the original versus perturbed forget data. As shown in \textbf{Fig.\,\ref{fig:overlap}(a)}, we use the overlap between these error sets as a proxy for measuring consistency in the forgetting targets. \textbf{Fig.\,\ref{fig:overlap}(b)} presents the overlap ratios across all perturbation types. Despite differences in format, such as random masking, semantic rewriting, and watermarking, all perturbed versions exhibit over 93\% overlap with the original unlearning. This high consistency suggests that semantically faithful perturbations preserve the key content needed for effective unlearning, even when the surface form of the data is significantly altered.

\begin{figure}[htb] 
\vspace*{0mm}
\centering  
\begin{tabular}{cc}
\hspace*{-2mm}
\includegraphics[width=0.23\textwidth]{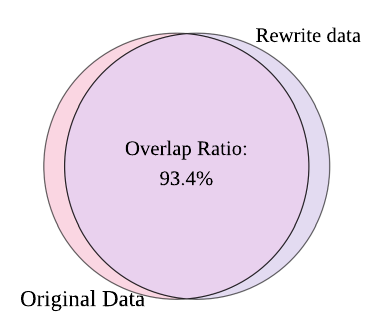} &
\hspace*{-4mm}\includegraphics[width=0.23\textwidth]{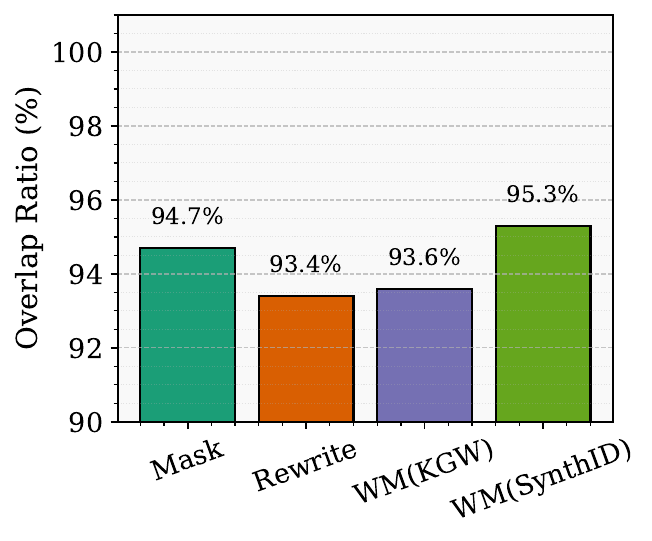}\\
  {\hspace*{-4mm} \scriptsize (a) Error Set Overlap}  &
  {\hspace*{-4mm} \scriptsize (b) Overlap Ratio} 
\end{tabular}
\vspace*{-3mm}
\caption{
Consistency of unlearning error rates under perturbed forget data.
(a) Venn diagram showing the overlap in incorrectly answered WMDP questions between models unlearned with original and rewritten forget data.
(b) Overlap ratios between the error sets of models unlearned with various perturbed forget sets, including Mask, Rewrite, WM(KGW), and WM(SynthID), and the baseline model trained with the original forget data. 
}
\label{fig:overlap}
\end{figure}


\begin{table*}[htb]
\begin{center}
\caption{\small{
Evaluation of unlearned models on the WMDP-Bio sentence completion task using different types of forget data.
Each row indicates models trained with original or perturbed forget data, where inputs are shown with highlighted \colorbox{LightCyan}{salient tokens} that contribute most to the forgetting objective. Model generation outputs are displayed to illustrate behavioral differences across unlearned models trained on different noisy forget sets. To quantify semantic preservation, overlap ratios of salient tokens are reported between each perturbed input and its original counterpart.
}}
\vspace*{-3mm}
\resizebox{\textwidth}{!}{
\begin{tabular}{c|c|c|c}
\toprule[1pt]
\textbf{Data Type} & 
\textbf{Forget Data Sample w/ Extracted \colorbox{LightCyan}{Salient Tokens}} & 
\textbf{Unlearned Model Generation} & 
\begin{tabular}{c}
\textbf{Salient Tokens} \\
\textbf{Overlap Ratio}
\end{tabular} \\
\midrule
\textbf{Original Data} & 
\begin{tabular}{c}

Since their first use as expression \colorbox{LightCyan}{vectors} in the 1980s, \colorbox{LightCyan}{Ad} \colorbox{LightCyan}{vectors} have received 
tremendous attention as \colorbox{LightCyan}{gene} delivery vehicles for \colorbox{LightCyan}{vaccine} \colorbox{LightCyan}{antigens}. They \\ have been 
extensively tested as \colorbox{LightCyan}{vaccine} delivery systems in several pre-\colorbox{LightCyan}{clinical} and 
\colorbox{LightCyan}{clinical}
studies for a number of \colorbox{LightCyan}{infectious} \colorbox{LightCyan}{diseases} including \colorbox{LightCyan}{measles}, \\
\colorbox{LightCyan}{hepatitis-B}, \colorbox{LightCyan}{rabies}, \colorbox{LightCyan}{anthrax}, \colorbox{LightCyan}{Ebola}, severe \colorbox{LightCyan}{acute} \colorbox{LightCyan}{respiratory} \colorbox{LightCyan}{syndrome}
(\colorbox{LightCyan}{SARS}), human \colorbox{LightCyan}{immunodeficiency} \colorbox{LightCyan}{virus} 1 (\colorbox{LightCyan}{HIV-1}), \colorbox{LightCyan}{malaria}, \\ \colorbox{LightCyan}{tuberculosis}, 
and \colorbox{LightCyan}{influenza}. There are two basic types of \colorbox{LightCyan}{Ad} \colorbox{LightCyan}{vectors} that are being
utilized for 
\colorbox{LightCyan}{gene} delivery applications. The first type of \colorbox{LightCyan}{Ad} \colorbox{LightCyan}{vectors},\underline{\hspace*{5mm}} 

\end{tabular} & 
\begin{tabular}{c}
.@the...\textbackslash{n}\textbackslash{n}\#\#\#.more\textbackslash{n}\textbackslash{n}.\textbackslash{n}\textbackslash{n}\#\#\#.s\textbackslash{n}\\
vin.x 
\textbackslash{n}.\textbackslash{n}\textbackslash{n}\#\#,\textbackslash{n}inway.\textbackslash{n}x.\textbackslash{n}\\
\textbackslash{n}@\textbackslash{n}@@
in.\textbackslash{n}
\#\#\#\textbackslash{n}xer\textbackslash{n}more\textbackslash{n}\textbackslash{n}b\textbackslash{n}\\
no\textbackslash{n}\textbackslash{n}w
\textbackslash{n}\textbackslash{n}There\textbackslash{n}@x\textbackslash{n}h\textbackslash{n}\textbackslash{n}\\
no\textbackslash{n}\textbackslash{n}
scundo\textbackslash{n}there.\textbackslash{n}\textbackslash{n}no\textbackslash{n}\textbackslash{n}\\
how.\textbackslash{n}\textbackslash{n}
Cural.\textbackslash{n}sair\textbackslash{n}\textbackslash{n}hg\textbackslash{n}.\textbackslash{n}'
\end{tabular} & 
\begin{tabular}{c} 100\% \end{tabular} \\
\midrule
\textbf{Mask Data} & 
\begin{tabular}{c}
Since their first *** as expression \colorbox{LightCyan}{vectors} in the 1980s, \colorbox{LightCyan}{Ad} \colorbox{LightCyan}{vectors} have received 
tremendous attention as \colorbox{LightCyan}{gene} delivery vehicles for \colorbox{LightCyan}{vaccine} \colorbox{LightCyan}{antigens}. They \\ have been 
extensively tested as \colorbox{LightCyan}{vaccine} delivery systems in several pre-\colorbox{LightCyan}{clinical} and 
\colorbox{LightCyan}{clinical}
studies for a number of \colorbox{LightCyan}{infectious} \colorbox{LightCyan}{diseases} including \colorbox{LightCyan}{measles}, \\
\colorbox{LightCyan}{hepatitis-B}, **** , \colorbox{LightCyan}{anthrax}, \colorbox{LightCyan}{Ebola}, severe \colorbox{LightCyan}{acute} \colorbox{LightCyan}{respiratory} *** 
(\colorbox{LightCyan}{SARS}), human **** \colorbox{LightCyan}{virus} 1 (\colorbox{LightCyan}{HIV-1}), \colorbox{LightCyan}{malaria}, \\ \colorbox{LightCyan}{tuberculosis}, 
and \colorbox{LightCyan}{influenza}. There are two basic types of \colorbox{LightCyan}{Ad} \colorbox{LightCyan}{vectors} that are being
utilized for 
\colorbox{LightCyan}{gene} delivery applications. The first *** of \colorbox{LightCyan}{Ad} \colorbox{LightCyan}{vectors},\underline{\hspace*{5mm}} 
\end{tabular} & 
\begin{tabular}{c}
\textbackslash{n}-*-\#-* 
\textbackslash{n}*---*- \\
\textbackslash{n}*\#-*\# 
\textbackslash{n}————— 
\textbackslash{n}*-* \\
\textbackslash{n}-\#-*-- 
\textbackslash{n}\#\#* 
\textbackslash{n}--* 
\textbackslash{n}\#** \\
\textbackslash{n}\#\#\#--* 
\textbackslash{n}**\# 
\textbackslash{n}\#--\#* 
\textbackslash{n}\#\#-* 
\textbackslash{n}-*\#\# \\
\textbackslash{n}--\#-* 
\textbackslash{n}-****- 
\textbackslash{n}--\#\#- 
\textbackslash{n}*\#**\#* 
\end{tabular}
 & 
\begin{tabular}{c} 89.4\% \end{tabular} \\

\midrule
\textbf{Rewrite Data} & 
\begin{tabular}{c}
\colorbox{LightCyan}{Ad} \colorbox{LightCyan}{vectors}, first introduced in the 1980s as expression \colorbox{LightCyan}{vectors}, have since become a major 
focus of research for delivering \colorbox{LightCyan}{vaccine} \colorbox{LightCyan}{antigens} via \colorbox{LightCyan}{gene} transfer. \\ Their use has been 
extensively explored in both pre-\colorbox{LightCyan}{clinical} and \colorbox{LightCyan}{clinical} studies 
 targeting numerous \colorbox{LightCyan}{infectious} \colorbox{LightCyan}{diseases}, including \colorbox{LightCyan}{measles}, \colorbox{LightCyan}{hepatitis-B}, \colorbox{LightCyan}{rabies}, \\  \colorbox{LightCyan}{anthrax}, and \colorbox{LightCyan}{Ebola}. Other diseases such as severe \colorbox{LightCyan}{acute} \colorbox{LightCyan}{respiratory} \colorbox{LightCyan}{syndrome} (\colorbox{LightCyan}{SARS}), \colorbox{LightCyan}{human} \colorbox{LightCyan}{immunodeficiency} \colorbox{LightCyan}{virus} 1 (\colorbox{LightCyan}{HIV-1}), \colorbox{LightCyan}{malaria}, \colorbox{LightCyan}{tuberculosis}, and \\ \colorbox{LightCyan}{influenza} have also been the focus of such efforts. Currently, two major types of  \colorbox{LightCyan}{Ad} \colorbox{LightCyan}{vectors} are in use for \colorbox{LightCyan}{gene} delivery. Among them, the first type,\underline{\hspace*{5mm}}
\end{tabular} & 
\begin{tabular}{c}
\textbackslash{n}\#\#\#\textbackslash{n}\*\textbackslash{n}\*\#\textbackslash{n}\#\*\#\textbackslash{n}\#\#\*\textbackslash{n}\\
\textbackslash{n}\#\*\textbackslash{n}\#\textbackslash{n}\*\*\#\textbackslash{n}\#\textbackslash{n}\*\textbackslash{n}\*\#\#\\
\textbackslash{n}\*\*\*\textbackslash{n}\#\#\textbackslash{n}\#\*\*\#\textbackslash{n}\*\textbackslash{n}\#\*\textbackslash{n}\\
\textbackslash{n}\#\#\#\#\textbackslash{n}\*\#\textbackslash{n}\*\#\#\textbackslash{n}\#\*\*\textbackslash{n}\\
\textbackslash{n}\#\*\*\textbackslash{n}\#\#\textbackslash{n}\#\textbackslash{n}\*\*\#\textbackslash{n}

\end{tabular} & 
\begin{tabular}{c} 94.5\% \end{tabular} \\

\midrule
\textbf{WM(KGW) Data} & 
\begin{tabular}{c}
Since their introduction in the 1980s, \colorbox{LightCyan}{Ad} \colorbox{LightCyan}{vectors}—originally developed as expression \colorbox{LightCyan}{vectors}—have emerged as a central platform for delivering \colorbox{LightCyan}{vaccine} \colorbox{LightCyan}{antigens} \\ through \colorbox{LightCyan}{gene} transfer.  Over time, researchers have extensively explored their application in both pre-\colorbox{LightCyan}{clinical} and \colorbox{LightCyan}{clinical} contexts. These studies \\ have targeted a wide spectrum of \colorbox{LightCyan}{infectious} \colorbox{LightCyan}{diseases}, ranging from \colorbox{LightCyan}{measles}, \colorbox{LightCyan}{hepatitis-B}, and \colorbox{LightCyan}{rabies} to \colorbox{LightCyan}{anthrax} and \colorbox{LightCyan}{Ebola}. Additional pathogens \\ under investigation include those responsible for severe \colorbox{LightCyan}{acute} \colorbox{LightCyan}{respiratory} \colorbox{LightCyan}{syndrome} (\colorbox{LightCyan}{SARS}), \colorbox{LightCyan}{human} \colorbox{LightCyan}{immunodeficiency} \colorbox{LightCyan}{virus} 1 (\colorbox{LightCyan}{HIV-1}), \colorbox{LightCyan}{malaria}, \\ \colorbox{LightCyan}{tuberculosis}, and \colorbox{LightCyan}{influenza}. Presently, two primary categories of \colorbox{LightCyan}{Ad} \colorbox{LightCyan}{vectors} are employed in \colorbox{LightCyan}{gene} delivery strategies, with the first type described as follows:\underline{\hspace*{5mm}}
\end{tabular} & 
\begin{tabular}{c}

\textbackslash{n}\#\#\# 
\textbackslash{n}==* \
\textbackslash{n}*\# 
\textbackslash{n}\#*\# \\
\textbackslash{n}\#\#* 
\textbackslash{n}\#* 
\textbackslash{n}\# 
\textbackslash{n}**\# \\
\textbackslash{n}\#== 
\textbackslash{n}==* 
\textbackslash{n}*\#\# 
\textbackslash{n}*** \\
\textbackslash{n}==\#\# 
\textbackslash{n}\#==**\# 
\textbackslash{n}* 
\textbackslash{n}\#* \\
\textbackslash{n}\#\#\#\# 

\end{tabular} & 
\begin{tabular}{c} 92.3\% \end{tabular} \\

\bottomrule
\end{tabular}
}
\label{tab: keyword}
\end{center}
\end{table*}

\subsubsection{Saliency-based explanation for perturbation resilience.}
To understand why different perturbation strategies (Mask, Rewrite, and Watermark) do not significantly degrade unlearning performance, we conduct a saliency-based analysis. Using an LLM-as-a-judge framework, we automatically identify salient tokens in the original forget data, those that the LLM deems most critical for unlearning due to their biosecurity relevance (see \textbf{Appendix\,\ref{appendix:keywords}} for details).

For each perturbed forget set, we compute the proportion of these salient tokens that are preserved. As shown in \textbf{Table\,\ref{tab: keyword}}, even under rewriting or watermarking (via KGW), the majority of salient tokens remain intact (\textit{e.g.}, 94.5\% overlap in the rewrite case), indicating strong semantic retention.
To further validate this explanation, we compare unlearning performance using the full forget data versus only the extracted salient tokens. \textbf{Fig.\,\ref{fig:unlearn_saliency}} shows that unlearning with salient tokens alone achieves comparable efficacy to using the full data across all perturbation types. This confirms that unlearning robustness stems from the preservation of core saliency tokens, which retain the essential forgetting signal despite surface-level perturbations.

\begin{figure}[htb] 
\centering  
\includegraphics[width=0.4\textwidth]{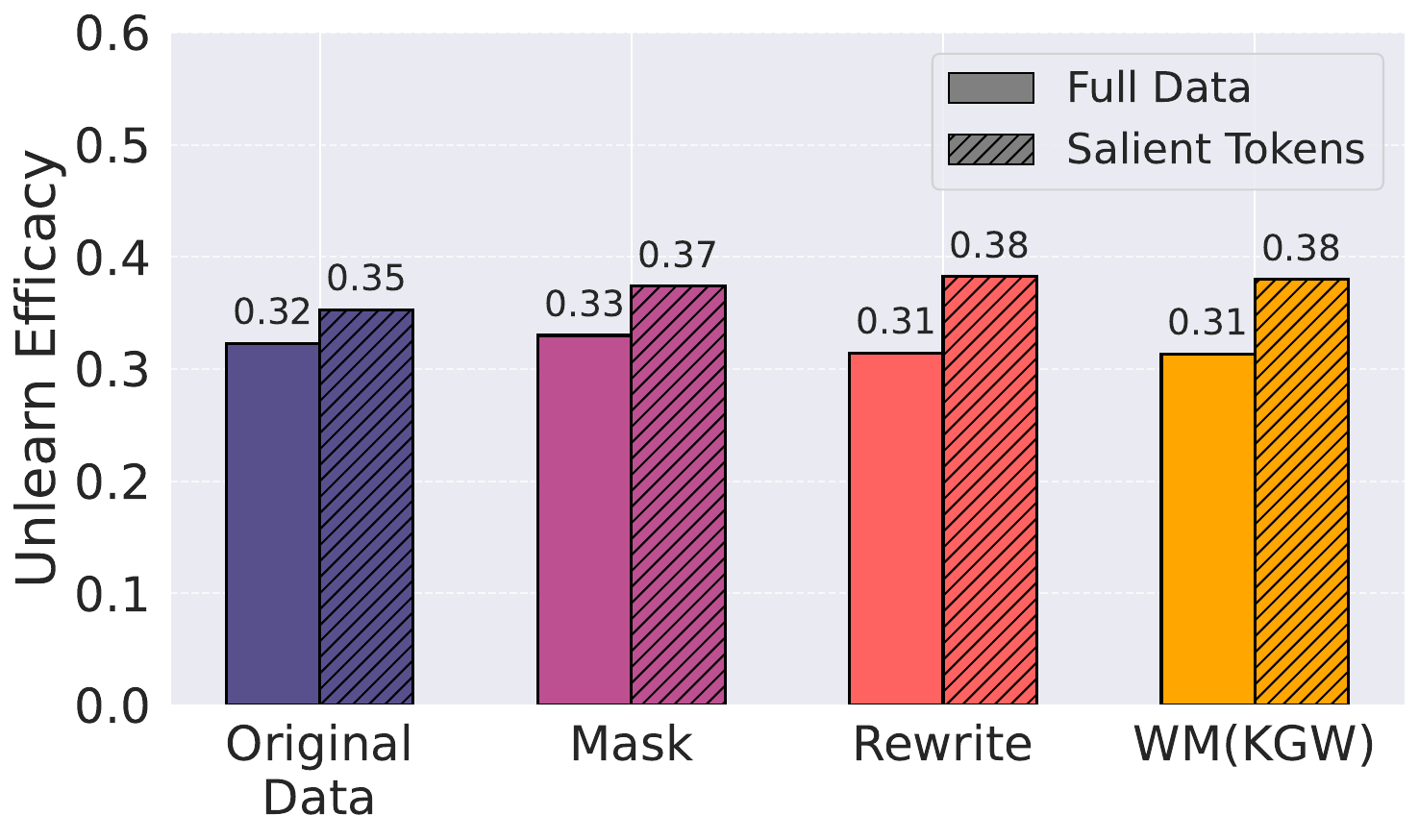} 
\caption{
Comparison of full data and salient token unlearning performance across different forget data types. This figure presents the unlearning efficacy of RMU on the WMDP dataset across three forget data perturbation types: Original Data, Mask, Rewrite, and WM(KGW). For each type, two variants are evaluated: using the entire forget set (Full Data) and using only LLM-as-judge-selected salient tokens (Salient Tokens). Results demonstrate that unlearning with salient tokens achieves efficacy comparable to Full Data unlearning across all settings, highlighting that a small, targeted subset of tokens is sufficient for effective unlearning when guided by LLM-based saliency selection.
}
\label{fig:unlearn_saliency}
\end{figure}

\section{Conclusion}
\label{sec: conclusion}


This work presents the first systematic analysis of how perturbed forget data—such as paraphrased rewrites, truncated or masked inputs, and watermarked synthetic content—impacts the performance of LLM unlearning. Despite substantial surface-level variations, we find that state-of-the-art methods like RMU and NPO remain surprisingly robust, with core semantic signals preserved and forgetting efficacy largely maintained across perturbation types. This resilience suggests that unlearning may depend less on the quantity and surface fidelity of the forget data, and more on the presence of essential semantic cues. These findings open a promising direction for attributing unlearning effectiveness to key data components and reinforce the value of a data-centric perspective in designing more efficient, interpretable, and practical unlearning systems.

\section{Limitations}

While this work presents the first systematic analysis of LLM unlearning under noisy forget data, there are still several limitations. First, our study is restricted to a few perturbation types—masking, rewriting, and watermarking—and two representative unlearning methods (RMU and NPO). The extent to which our findings generalize to other perturbations (\textit{e.g.}, adversarial corruptions) or unlearning techniques (\textit{e.g.}, targeted model editing or neuron-level interventions) remains unexplored. Second, our saliency-based interpretation relies on token-level semantic attribution, which may oversimplify the true mechanisms of forgetting. Future work could leverage causal or representation-level analyses to uncover deeper insights into unlearning robustness under data noise.

\section{Broader Impact}

This work provides the first empirical evidence that LLM unlearning can remain effective even when forget data is noisy, incomplete, or watermarked. This robustness has promising implications for real-world deployment, where clean and well-curated forget datasets are often unavailable. Our findings suggest that user-driven or regulatory data removal requests may still be reliably addressed despite imperfections in the forget data. However, this tolerance also introduces potential risks: adversaries could exploit it by submitting low-fidelity or adversarial forget requests to degrade or manipulate model behavior. Additionally, as synthetic and watermarked data become more prevalent in training pipelines, understanding how such signals interact with unlearning remains an important ethical and technical challenge. Ensuring secure and responsible deployment of unlearning systems will require careful consideration of these dynamics.

\begin{acks}
The work of C. Wang, Y. Zhang, J. Jia, and S. Liu was supported in part by the  NSF CISE Core Program Awards IIS-2207052 and IIS-2504263, the NSF CAREER Award IIS-2338068, the Amazon Faculty Research Award in AI for Information Security, the Cisco Faculty Research Award, the Open Philanthropy Research Grant, and the Center of AI Safety Research. 
\end{acks}

\bibliographystyle{ACM-Reference-Format}

\bibliography{acmart}


\begin{thebibliography}{60}


\ifx \showCODEN    \undefined \def \showCODEN     #1{\unskip}     \fi
\ifx \showISBNx    \undefined \def \showISBNx     #1{\unskip}     \fi
\ifx \showISBNxiii \undefined \def \showISBNxiii  #1{\unskip}     \fi
\ifx \showISSN     \undefined \def \showISSN      #1{\unskip}     \fi
\ifx \showLCCN     \undefined \def \showLCCN      #1{\unskip}     \fi
\ifx \shownote     \undefined \def \shownote      #1{#1}          \fi
\ifx \showarticletitle \undefined \def \showarticletitle #1{#1}   \fi
\ifx \showURL      \undefined \def \showURL       {\relax}        \fi
\providecommand\bibfield[2]{#2}
\providecommand\bibinfo[2]{#2}
\providecommand\natexlab[1]{#1}
\providecommand\showeprint[2][]{arXiv:#2}

\bibitem[Achiam et~al\mbox{.}(2023)]%
        {achiam2023gpt}
\bibfield{author}{\bibinfo{person}{Josh Achiam}, \bibinfo{person}{Steven Adler}, \bibinfo{person}{Sandhini Agarwal}, \bibinfo{person}{Lama Ahmad}, \bibinfo{person}{Ilge Akkaya}, \bibinfo{person}{Florencia~Leoni Aleman}, \bibinfo{person}{Diogo Almeida}, \bibinfo{person}{Janko Altenschmidt}, \bibinfo{person}{Sam Altman}, \bibinfo{person}{Shyamal Anadkat}, {et~al\mbox{.}}} \bibinfo{year}{2023}\natexlab{}.
\newblock \showarticletitle{Gpt-4 technical report}.
\newblock \bibinfo{journal}{\emph{arXiv preprint arXiv:2303.08774}} (\bibinfo{year}{2023}).
\newblock


\bibitem[Barbulescu and Triantafillou(2024)]%
        {barbulescu2024each}
\bibfield{author}{\bibinfo{person}{George-Octavian Barbulescu} {and} \bibinfo{person}{Peter Triantafillou}.} \bibinfo{year}{2024}\natexlab{}.
\newblock \showarticletitle{To Each (Textual Sequence) Its Own: Improving Memorized-Data Unlearning in Large Language Models}.
\newblock \bibinfo{journal}{\emph{arXiv preprint arXiv:2405.03097}} (\bibinfo{year}{2024}).
\newblock


\bibitem[Chen et~al\mbox{.}(2025)]%
        {chen2025safety}
\bibfield{author}{\bibinfo{person}{Yiwei Chen}, \bibinfo{person}{Yuguang Yao}, \bibinfo{person}{Yihua Zhang}, \bibinfo{person}{Bingquan Shen}, \bibinfo{person}{Gaowen Liu}, {and} \bibinfo{person}{Sijia Liu}.} \bibinfo{year}{2025}\natexlab{}.
\newblock \showarticletitle{Safety mirage: How spurious correlations undermine vlm safety fine-tuning}.
\newblock \bibinfo{journal}{\emph{arXiv preprint arXiv:2503.11832}} (\bibinfo{year}{2025}).
\newblock


\bibitem[Dathathri et~al\mbox{.}(2024)]%
        {dathathri2024scalable}
\bibfield{author}{\bibinfo{person}{Sumanth Dathathri}, \bibinfo{person}{Abigail See}, \bibinfo{person}{Sumedh Ghaisas}, \bibinfo{person}{Po-Sen Huang}, \bibinfo{person}{Rob McAdam}, \bibinfo{person}{Johannes Welbl}, \bibinfo{person}{Vandana Bachani}, \bibinfo{person}{Alex Kaskasoli}, \bibinfo{person}{Robert Stanforth}, \bibinfo{person}{Tatiana Matejovicova}, {et~al\mbox{.}}} \bibinfo{year}{2024}\natexlab{}.
\newblock \showarticletitle{Scalable watermarking for identifying large language model outputs}.
\newblock \bibinfo{journal}{\emph{Nature}} \bibinfo{volume}{634}, \bibinfo{number}{8035} (\bibinfo{year}{2024}), \bibinfo{pages}{818--823}.
\newblock


\bibitem[Deeb and Roger(2024)]%
        {deeb2024unlearning}
\bibfield{author}{\bibinfo{person}{Aghyad Deeb} {and} \bibinfo{person}{Fabien Roger}.} \bibinfo{year}{2024}\natexlab{}.
\newblock \showarticletitle{Do Unlearning Methods Remove Information from Language Model Weights?}
\newblock \bibinfo{journal}{\emph{arXiv preprint arXiv:2410.08827}} (\bibinfo{year}{2024}).
\newblock


\bibitem[Eldan and Russinovich(2023)]%
        {eldan2023whos}
\bibfield{author}{\bibinfo{person}{Ronen Eldan} {and} \bibinfo{person}{Mark Russinovich}.} \bibinfo{year}{2023}\natexlab{}.
\newblock \showarticletitle{Who's Harry Potter? Approximate Unlearning in LLMs}.
\newblock \bibinfo{journal}{\emph{arXiv preprint arXiv:2310.02238}} (\bibinfo{year}{2023}).
\newblock


\bibitem[Goldblum et~al\mbox{.}(2022)]%
        {goldblum2022dataset}
\bibfield{author}{\bibinfo{person}{Micah Goldblum}, \bibinfo{person}{Dimitris Tsipras}, \bibinfo{person}{Chulin Xie}, \bibinfo{person}{Xinyun Chen}, \bibinfo{person}{Avi Schwarzschild}, \bibinfo{person}{Dawn Song}, \bibinfo{person}{Aleksander M{\k{a}}dry}, \bibinfo{person}{Bo Li}, {and} \bibinfo{person}{Tom Goldstein}.} \bibinfo{year}{2022}\natexlab{}.
\newblock \showarticletitle{Dataset security for machine learning: Data poisoning, backdoor attacks, and defenses}.
\newblock \bibinfo{journal}{\emph{IEEE Transactions on Pattern Analysis and Machine Intelligence}} \bibinfo{volume}{45}, \bibinfo{number}{2} (\bibinfo{year}{2022}), \bibinfo{pages}{1563--1580}.
\newblock


\bibitem[Hendrycks et~al\mbox{.}(2020)]%
        {hendrycks2020measuring}
\bibfield{author}{\bibinfo{person}{Dan Hendrycks}, \bibinfo{person}{Collin Burns}, \bibinfo{person}{Steven Basart}, \bibinfo{person}{Andy Zou}, \bibinfo{person}{Mantas Mazeika}, \bibinfo{person}{Dawn Song}, {and} \bibinfo{person}{Jacob Steinhardt}.} \bibinfo{year}{2020}\natexlab{}.
\newblock \showarticletitle{Measuring massive multitask language understanding}.
\newblock \bibinfo{journal}{\emph{arXiv preprint arXiv:2009.03300}} (\bibinfo{year}{2020}).
\newblock


\bibitem[Hu et~al\mbox{.}(2024)]%
        {hu2024jogging}
\bibfield{author}{\bibinfo{person}{Shengyuan Hu}, \bibinfo{person}{Yiwei Fu}, \bibinfo{person}{Zhiwei~Steven Wu}, {and} \bibinfo{person}{Virginia Smith}.} \bibinfo{year}{2024}\natexlab{}.
\newblock \showarticletitle{Jogging the Memory of Unlearned Model Through Targeted Relearning Attack}.
\newblock \bibinfo{journal}{\emph{arXiv preprint arXiv:2406.13356}} (\bibinfo{year}{2024}).
\newblock


\bibitem[Hu et~al\mbox{.}(2023)]%
        {hu2023unbiased}
\bibfield{author}{\bibinfo{person}{Zhengmian Hu}, \bibinfo{person}{Lichang Chen}, \bibinfo{person}{Xidong Wu}, \bibinfo{person}{Yihan Wu}, \bibinfo{person}{Hongyang Zhang}, {and} \bibinfo{person}{Heng Huang}.} \bibinfo{year}{2023}\natexlab{}.
\newblock \showarticletitle{Unbiased watermark for large language models}.
\newblock \bibinfo{journal}{\emph{arXiv preprint arXiv:2310.10669}} (\bibinfo{year}{2023}).
\newblock


\bibitem[Huang et~al\mbox{.}(2024)]%
        {sun2024trustllm}
\bibfield{author}{\bibinfo{person}{Yue Huang}, \bibinfo{person}{Lichao Sun}, \bibinfo{person}{Haoran Wang}, \bibinfo{person}{Siyuan Wu}, \bibinfo{person}{Qihui Zhang}, \bibinfo{person}{Yuan Li}, \bibinfo{person}{Chujie Gao}, \bibinfo{person}{Yixin Huang}, {et~al\mbox{.}}} \bibinfo{year}{2024}\natexlab{}.
\newblock \showarticletitle{Position: {T}rust{LLM}: Trustworthiness in Large Language Models}. In \bibinfo{booktitle}{\emph{Proceedings of the 41st International Conference on Machine Learning}} \emph{(\bibinfo{series}{Proceedings of Machine Learning Research}, Vol.~\bibinfo{volume}{235})}. \bibinfo{pages}{20166--20270}.
\newblock


\bibitem[Ilharco et~al\mbox{.}(2023)]%
        {ilharco2022editing}
\bibfield{author}{\bibinfo{person}{Gabriel Ilharco}, \bibinfo{person}{Marco~Tulio Ribeiro}, \bibinfo{person}{Mitchell Wortsman}, \bibinfo{person}{Ludwig Schmidt}, \bibinfo{person}{Hannaneh Hajishirzi}, {and} \bibinfo{person}{Ali Farhadi}.} \bibinfo{year}{2023}\natexlab{}.
\newblock \showarticletitle{Editing models with task arithmetic}. In \bibinfo{booktitle}{\emph{The Eleventh International Conference on Learning Representations}}.
\newblock


\bibitem[Jang et~al\mbox{.}(2023)]%
        {jang2022knowledge}
\bibfield{author}{\bibinfo{person}{Joel Jang}, \bibinfo{person}{Dongkeun Yoon}, \bibinfo{person}{Sohee Yang}, \bibinfo{person}{Sungmin Cha}, \bibinfo{person}{Moontae Lee}, \bibinfo{person}{Lajanugen Logeswaran}, {and} \bibinfo{person}{Minjoon Seo}.} \bibinfo{year}{2023}\natexlab{}.
\newblock \showarticletitle{Knowledge Unlearning for Mitigating Privacy Risks in Language Models}. In \bibinfo{booktitle}{\emph{Proceedings of the 61st Annual Meeting of the Association for Computational Linguistics}}. \bibinfo{publisher}{Association for Computational Linguistics}, \bibinfo{pages}{14389--14408}.
\newblock


\bibitem[Jia et~al\mbox{.}(2024)]%
        {jia2024wagle}
\bibfield{author}{\bibinfo{person}{Jinghan Jia}, \bibinfo{person}{Jiancheng Liu}, \bibinfo{person}{Yihua Zhang}, \bibinfo{person}{Parikshit Ram}, \bibinfo{person}{Nathalie Baracaldo}, {and} \bibinfo{person}{Sijia Liu}.} \bibinfo{year}{2024}\natexlab{}.
\newblock \showarticletitle{{WAGLE}: Strategic Weight Attribution for Effective and Modular Unlearning in Large Language Models}. In \bibinfo{booktitle}{\emph{The Thirty-eighth Annual Conference on Neural Information Processing Systems}}.
\newblock


\bibitem[Kirchenbauer et~al\mbox{.}(2023)]%
        {kirchenbauer2023watermark}
\bibfield{author}{\bibinfo{person}{John Kirchenbauer}, \bibinfo{person}{Jonas Geiping}, \bibinfo{person}{Yuxin Wen}, \bibinfo{person}{Jonathan Katz}, \bibinfo{person}{Ian Miers}, {and} \bibinfo{person}{Tom Goldstein}.} \bibinfo{year}{2023}\natexlab{}.
\newblock \showarticletitle{A watermark for large language models}. In \bibinfo{booktitle}{\emph{International Conference on Machine Learning}}. PMLR, \bibinfo{pages}{17061--17084}.
\newblock


\bibitem[Lee et~al\mbox{.}(2023)]%
        {lee2023wrote}
\bibfield{author}{\bibinfo{person}{Taehyun Lee}, \bibinfo{person}{Seokhee Hong}, \bibinfo{person}{Jaewoo Ahn}, \bibinfo{person}{Ilgee Hong}, \bibinfo{person}{Hwaran Lee}, \bibinfo{person}{Sangdoo Yun}, \bibinfo{person}{Jamin Shin}, {and} \bibinfo{person}{Gunhee Kim}.} \bibinfo{year}{2023}\natexlab{}.
\newblock \showarticletitle{Who wrote this code? watermarking for code generation}.
\newblock \bibinfo{journal}{\emph{arXiv preprint arXiv:2305.15060}} (\bibinfo{year}{2023}).
\newblock


\bibitem[Li et~al\mbox{.}(2024b)]%
        {li2024wmdp}
\bibfield{author}{\bibinfo{person}{Nathaniel Li}, \bibinfo{person}{Alexander Pan}, \bibinfo{person}{Anjali Gopal}, \bibinfo{person}{Summer Yue}, \bibinfo{person}{Daniel Berrios}, \bibinfo{person}{Alice Gatti}, \bibinfo{person}{Justin~D. Li}, \bibinfo{person}{Ann-Kathrin Dombrowski}, \bibinfo{person}{Shashwat Goel}, \bibinfo{person}{Gabriel Mukobi}, \bibinfo{person}{Nathan Helm-Burger}, \bibinfo{person}{Rassin Lababidi}, \bibinfo{person}{Lennart Justen}, \bibinfo{person}{Andrew~Bo Liu}, \bibinfo{person}{Michael Chen}, \bibinfo{person}{Isabelle Barrass}, \bibinfo{person}{Oliver Zhang}, \bibinfo{person}{Xiaoyuan Zhu}, \bibinfo{person}{Rishub Tamirisa}, \bibinfo{person}{Bhrugu Bharathi}, \bibinfo{person}{Ariel Herbert-Voss}, \bibinfo{person}{Cort~B Breuer}, \bibinfo{person}{Andy Zou}, \bibinfo{person}{Mantas Mazeika}, \bibinfo{person}{Zifan Wang}, \bibinfo{person}{Palash Oswal}, \bibinfo{person}{Weiran Lin}, \bibinfo{person}{Adam~Alfred Hunt}, \bibinfo{person}{Justin Tienken-Harder}, \bibinfo{person}{Kevin~Y.
  Shih}, \bibinfo{person}{Kemper Talley}, \bibinfo{person}{John Guan}, \bibinfo{person}{Ian Steneker}, \bibinfo{person}{David Campbell}, \bibinfo{person}{Brad Jokubaitis}, \bibinfo{person}{Steven Basart}, \bibinfo{person}{Stephen Fitz}, \bibinfo{person}{Ponnurangam Kumaraguru}, \bibinfo{person}{Kallol~Krishna Karmakar}, \bibinfo{person}{Uday Tupakula}, \bibinfo{person}{Vijay Varadharajan}, \bibinfo{person}{Yan Shoshitaishvili}, \bibinfo{person}{Jimmy Ba}, \bibinfo{person}{Kevin~M. Esvelt}, \bibinfo{person}{Alexandr Wang}, {and} \bibinfo{person}{Dan Hendrycks}.} \bibinfo{year}{2024}\natexlab{b}.
\newblock \showarticletitle{The {WMDP} Benchmark: Measuring and Reducing Malicious Use with Unlearning}. In \bibinfo{booktitle}{\emph{Proceedings of the 41st International Conference on Machine Learning}}. \bibinfo{pages}{28525--28550}.
\newblock


\bibitem[Li et~al\mbox{.}(2024a)]%
        {li2024data}
\bibfield{author}{\bibinfo{person}{Yinheng Li}, \bibinfo{person}{Rogerio Bonatti}, \bibinfo{person}{Sara Abdali}, \bibinfo{person}{Justin Wagle}, {and} \bibinfo{person}{Kazuhito Koishida}.} \bibinfo{year}{2024}\natexlab{a}.
\newblock \showarticletitle{Data generation using large language models for text classification: An empirical case study}.
\newblock \bibinfo{journal}{\emph{arXiv preprint arXiv:2407.12813}} (\bibinfo{year}{2024}).
\newblock


\bibitem[Liu et~al\mbox{.}(2024a)]%
        {liu2024deepseek}
\bibfield{author}{\bibinfo{person}{Aixin Liu}, \bibinfo{person}{Bei Feng}, \bibinfo{person}{Bin Wang}, \bibinfo{person}{Bingxuan Wang}, \bibinfo{person}{Bo Liu}, \bibinfo{person}{Chenggang Zhao}, \bibinfo{person}{Chengqi Dengr}, \bibinfo{person}{Chong Ruan}, \bibinfo{person}{Damai Dai}, \bibinfo{person}{Daya Guo}, {et~al\mbox{.}}} \bibinfo{year}{2024}\natexlab{a}.
\newblock \showarticletitle{Deepseek-v2: A strong, economical, and efficient mixture-of-experts language model}.
\newblock \bibinfo{journal}{\emph{arXiv preprint arXiv:2405.04434}} (\bibinfo{year}{2024}).
\newblock


\bibitem[Liu and Mozafari(2024)]%
        {liu2024query}
\bibfield{author}{\bibinfo{person}{Jie Liu} {and} \bibinfo{person}{Barzan Mozafari}.} \bibinfo{year}{2024}\natexlab{}.
\newblock \showarticletitle{Query rewriting via large language models}.
\newblock \bibinfo{journal}{\emph{arXiv preprint arXiv:2403.09060}} (\bibinfo{year}{2024}).
\newblock


\bibitem[Liu et~al\mbox{.}(2024b)]%
        {liu2024best}
\bibfield{author}{\bibinfo{person}{Ruibo Liu}, \bibinfo{person}{Jerry Wei}, \bibinfo{person}{Fangyu Liu}, \bibinfo{person}{Chenglei Si}, \bibinfo{person}{Yanzhe Zhang}, \bibinfo{person}{Jinmeng Rao}, \bibinfo{person}{Steven Zheng}, \bibinfo{person}{Daiyi Peng}, \bibinfo{person}{Diyi Yang}, \bibinfo{person}{Denny Zhou}, {et~al\mbox{.}}} \bibinfo{year}{2024}\natexlab{b}.
\newblock \showarticletitle{Best practices and lessons learned on synthetic data}.
\newblock \bibinfo{journal}{\emph{arXiv preprint arXiv:2404.07503}} (\bibinfo{year}{2024}).
\newblock


\bibitem[Liu et~al\mbox{.}(2024c)]%
        {liu2024rethinking}
\bibfield{author}{\bibinfo{person}{Sijia Liu}, \bibinfo{person}{Yuanshun Yao}, \bibinfo{person}{Jinghan Jia}, \bibinfo{person}{Stephen Casper}, \bibinfo{person}{Nathalie Baracaldo}, \bibinfo{person}{Peter Hase}, \bibinfo{person}{Yuguang Yao}, \bibinfo{person}{Chris~Yuhao Liu}, \bibinfo{person}{Xiaojun Xu}, \bibinfo{person}{Hang Li}, \bibinfo{person}{Kush~R. Varshney}, \bibinfo{person}{Mohit Bansal}, \bibinfo{person}{Sanmi Koyejo}, {and} \bibinfo{person}{Yang Liu}.} \bibinfo{year}{2024}\natexlab{c}.
\newblock \showarticletitle{Rethinking machine unlearning for large language models}.
\newblock \bibinfo{journal}{\emph{arXiv preprint arXiv:2402.08787}} (\bibinfo{year}{2024}).
\newblock


\bibitem[Lu et~al\mbox{.}(2022)]%
        {lu2022quark}
\bibfield{author}{\bibinfo{person}{Ximing Lu}, \bibinfo{person}{Sean Welleck}, \bibinfo{person}{Jack Hessel}, \bibinfo{person}{Liwei Jiang}, \bibinfo{person}{Lianhui Qin}, \bibinfo{person}{Peter West}, \bibinfo{person}{Prithviraj Ammanabrolu}, {and} \bibinfo{person}{Yejin Choi}.} \bibinfo{year}{2022}\natexlab{}.
\newblock \showarticletitle{Quark: Controllable text generation with reinforced unlearning}.
\newblock \bibinfo{journal}{\emph{Advances in neural information processing systems}}  \bibinfo{volume}{35} (\bibinfo{year}{2022}), \bibinfo{pages}{27591--27609}.
\newblock


\bibitem[{\L}ucki et~al\mbox{.}(2024)]%
        {lucki2024adversarial}
\bibfield{author}{\bibinfo{person}{Jakub {\L}ucki}, \bibinfo{person}{Boyi Wei}, \bibinfo{person}{Yangsibo Huang}, \bibinfo{person}{Peter Henderson}, \bibinfo{person}{Florian Tram{\`e}r}, {and} \bibinfo{person}{Javier Rando}.} \bibinfo{year}{2024}\natexlab{}.
\newblock \showarticletitle{An adversarial perspective on machine unlearning for ai safety}.
\newblock \bibinfo{journal}{\emph{arXiv preprint arXiv:2409.18025}} (\bibinfo{year}{2024}).
\newblock


\bibitem[Lupidi et~al\mbox{.}(2024)]%
        {lupidi2024source2synth}
\bibfield{author}{\bibinfo{person}{Alisia Lupidi}, \bibinfo{person}{Carlos Gemmell}, \bibinfo{person}{Nicola Cancedda}, \bibinfo{person}{Jane Dwivedi-Yu}, \bibinfo{person}{Jason Weston}, \bibinfo{person}{Jakob Foerster}, \bibinfo{person}{Roberta Raileanu}, {and} \bibinfo{person}{Maria Lomeli}.} \bibinfo{year}{2024}\natexlab{}.
\newblock \showarticletitle{Source2synth: Synthetic data generation and curation grounded in real data sources}.
\newblock \bibinfo{journal}{\emph{arXiv preprint arXiv:2409.08239}} (\bibinfo{year}{2024}).
\newblock


\bibitem[Lynch et~al\mbox{.}(2024)]%
        {lynch2024eight}
\bibfield{author}{\bibinfo{person}{Aengus Lynch}, \bibinfo{person}{Phillip Guo}, \bibinfo{person}{Aidan Ewart}, \bibinfo{person}{Stephen Casper}, {and} \bibinfo{person}{Dylan Hadfield-Menell}.} \bibinfo{year}{2024}\natexlab{}.
\newblock \showarticletitle{Eight methods to evaluate robust unlearning in llms}.
\newblock \bibinfo{journal}{\emph{arXiv preprint arXiv:2402.16835}} (\bibinfo{year}{2024}).
\newblock


\bibitem[Maini et~al\mbox{.}(2024)]%
        {maini2024tofu}
\bibfield{author}{\bibinfo{person}{Pratyush Maini}, \bibinfo{person}{Zhili Feng}, \bibinfo{person}{Avi Schwarzschild}, \bibinfo{person}{Zachary~Chase Lipton}, {and} \bibinfo{person}{J~Zico Kolter}.} \bibinfo{year}{2024}\natexlab{}.
\newblock \showarticletitle{{TOFU}: A Task of Fictitious Unlearning for {LLM}s}. In \bibinfo{booktitle}{\emph{First Conference on Language Modeling}}.
\newblock


\bibitem[Meng et~al\mbox{.}(2022)]%
        {meng2022locating}
\bibfield{author}{\bibinfo{person}{Kevin Meng}, \bibinfo{person}{David Bau}, \bibinfo{person}{Alex Andonian}, {and} \bibinfo{person}{Yonatan Belinkov}.} \bibinfo{year}{2022}\natexlab{}.
\newblock \showarticletitle{Locating and editing factual associations in GPT}.
\newblock \bibinfo{journal}{\emph{Advances in Neural Information Processing Systems}}  \bibinfo{volume}{35} (\bibinfo{year}{2022}), \bibinfo{pages}{17359--17372}.
\newblock


\bibitem[Merity et~al\mbox{.}(2016)]%
        {merity2016pointer}
\bibfield{author}{\bibinfo{person}{Stephen Merity}, \bibinfo{person}{Caiming Xiong}, \bibinfo{person}{James Bradbury}, {and} \bibinfo{person}{Richard Socher}.} \bibinfo{year}{2016}\natexlab{}.
\newblock \bibinfo{title}{Pointer Sentinel Mixture Models}.
\newblock
\showeprint[arxiv]{1609.07843}~[cs.CL]


\bibitem[Motoki et~al\mbox{.}(2023)]%
        {motoki2023more}
\bibfield{author}{\bibinfo{person}{Fabio Motoki}, \bibinfo{person}{Valdemar Pinho~Neto}, {and} \bibinfo{person}{Victor Rodrigues}.} \bibinfo{year}{2023}\natexlab{}.
\newblock \showarticletitle{More human than human: Measuring chatgpt political bias}.
\newblock \bibinfo{journal}{\emph{Available at SSRN 4372349}} (\bibinfo{year}{2023}).
\newblock


\bibitem[Pal et~al\mbox{.}(2025)]%
        {pal2025llm}
\bibfield{author}{\bibinfo{person}{Soumyadeep Pal}, \bibinfo{person}{Changsheng Wang}, \bibinfo{person}{James Diffenderfer}, \bibinfo{person}{Bhavya Kailkhura}, {and} \bibinfo{person}{Sijia Liu}.} \bibinfo{year}{2025}\natexlab{}.
\newblock \showarticletitle{Llm unlearning reveals a stronger-than-expected coreset effect in current benchmarks}.
\newblock \bibinfo{journal}{\emph{arXiv preprint arXiv:2504.10185}} (\bibinfo{year}{2025}).
\newblock


\bibitem[Patel et~al\mbox{.}(2024)]%
        {patel2024datadreamer}
\bibfield{author}{\bibinfo{person}{Ajay Patel}, \bibinfo{person}{Colin Raffel}, {and} \bibinfo{person}{Chris Callison-Burch}.} \bibinfo{year}{2024}\natexlab{}.
\newblock \showarticletitle{Datadreamer: A tool for synthetic data generation and reproducible llm workflows}.
\newblock \bibinfo{journal}{\emph{arXiv preprint arXiv:2402.10379}} (\bibinfo{year}{2024}).
\newblock


\bibitem[Patil et~al\mbox{.}(2024)]%
        {patil2023can}
\bibfield{author}{\bibinfo{person}{Vaidehi Patil}, \bibinfo{person}{Peter Hase}, {and} \bibinfo{person}{Mohit Bansal}.} \bibinfo{year}{2024}\natexlab{}.
\newblock \showarticletitle{Can Sensitive Information Be Deleted From LLMs? Objectives for Defending Against Extraction Attacks}.
\newblock \bibinfo{journal}{\emph{ICLR}} (\bibinfo{year}{2024}).
\newblock


\bibitem[Patil et~al\mbox{.}(2025)]%
        {patil2025upcore}
\bibfield{author}{\bibinfo{person}{Vaidehi Patil}, \bibinfo{person}{Elias Stengel-Eskin}, {and} \bibinfo{person}{Mohit Bansal}.} \bibinfo{year}{2025}\natexlab{}.
\newblock \showarticletitle{Upcore: Utility-preserving coreset selection for balanced unlearning}.
\newblock \bibinfo{journal}{\emph{arXiv preprint arXiv:2502.15082}} (\bibinfo{year}{2025}).
\newblock


\bibitem[Pawelczyk et~al\mbox{.}(2023)]%
        {pawelczyk2023context}
\bibfield{author}{\bibinfo{person}{Martin Pawelczyk}, \bibinfo{person}{Seth Neel}, {and} \bibinfo{person}{Himabindu Lakkaraju}.} \bibinfo{year}{2023}\natexlab{}.
\newblock \showarticletitle{In-context unlearning: Language models as few shot unlearners}.
\newblock \bibinfo{journal}{\emph{arXiv preprint arXiv:2310.07579}} (\bibinfo{year}{2023}).
\newblock


\bibitem[Rafailov et~al\mbox{.}(2024)]%
        {rafailov2024direct}
\bibfield{author}{\bibinfo{person}{Rafael Rafailov}, \bibinfo{person}{Archit Sharma}, \bibinfo{person}{Eric Mitchell}, \bibinfo{person}{Christopher~D Manning}, \bibinfo{person}{Stefano Ermon}, {and} \bibinfo{person}{Chelsea Finn}.} \bibinfo{year}{2024}\natexlab{}.
\newblock \showarticletitle{Direct preference optimization: Your language model is secretly a reward model}.
\newblock \bibinfo{journal}{\emph{Advances in Neural Information Processing Systems}}  \bibinfo{volume}{36} (\bibinfo{year}{2024}).
\newblock


\bibitem[Schoepf et~al\mbox{.}(2025)]%
        {schoepf2025redirection}
\bibfield{author}{\bibinfo{person}{Stefan Schoepf}, \bibinfo{person}{Michael~Curtis Mozer}, \bibinfo{person}{Nicole~Elyse Mitchell}, \bibinfo{person}{Alexandra Brintrup}, \bibinfo{person}{Georgios Kaissis}, \bibinfo{person}{Peter Kairouz}, {and} \bibinfo{person}{Eleni Triantafillou}.} \bibinfo{year}{2025}\natexlab{}.
\newblock \showarticletitle{Redirection for Erasing Memory (REM): Towards a universal unlearning method for corrupted data}.
\newblock \bibinfo{journal}{\emph{arXiv preprint arXiv:2505.17730}} (\bibinfo{year}{2025}).
\newblock


\bibitem[Sennrich et~al\mbox{.}(2015)]%
        {sennrich2015improving}
\bibfield{author}{\bibinfo{person}{Rico Sennrich}, \bibinfo{person}{Barry Haddow}, {and} \bibinfo{person}{Alexandra Birch}.} \bibinfo{year}{2015}\natexlab{}.
\newblock \showarticletitle{Improving neural machine translation models with monolingual data}.
\newblock \bibinfo{journal}{\emph{arXiv preprint arXiv:1511.06709}} (\bibinfo{year}{2015}).
\newblock


\bibitem[Shi et~al\mbox{.}(2024)]%
        {shi2024muse}
\bibfield{author}{\bibinfo{person}{Weijia Shi}, \bibinfo{person}{Jaechan Lee}, \bibinfo{person}{Yangsibo Huang}, \bibinfo{person}{Sadhika Malladi}, \bibinfo{person}{Jieyu Zhao}, \bibinfo{person}{Ari Holtzman}, \bibinfo{person}{Daogao Liu}, \bibinfo{person}{Luke Zettlemoyer}, \bibinfo{person}{Noah~A Smith}, {and} \bibinfo{person}{Chiyuan Zhang}.} \bibinfo{year}{2024}\natexlab{}.
\newblock \showarticletitle{Muse: Machine unlearning six-way evaluation for language models}.
\newblock \bibinfo{journal}{\emph{arXiv preprint arXiv:2407.06460}} (\bibinfo{year}{2024}).
\newblock


\bibitem[Shu et~al\mbox{.}(2024)]%
        {shu2024rewritelm}
\bibfield{author}{\bibinfo{person}{Lei Shu}, \bibinfo{person}{Liangchen Luo}, \bibinfo{person}{Jayakumar Hoskere}, \bibinfo{person}{Yun Zhu}, \bibinfo{person}{Yinxiao Liu}, \bibinfo{person}{Simon Tong}, \bibinfo{person}{Jindong Chen}, {and} \bibinfo{person}{Lei Meng}.} \bibinfo{year}{2024}\natexlab{}.
\newblock \showarticletitle{Rewritelm: An instruction-tuned large language model for text rewriting}. In \bibinfo{booktitle}{\emph{Proceedings of the AAAI Conference on Artificial Intelligence}}, Vol.~\bibinfo{volume}{38}. \bibinfo{pages}{18970--18980}.
\newblock


\bibitem[Sun et~al\mbox{.}(2024a)]%
        {sun2024forget}
\bibfield{author}{\bibinfo{person}{Changchang Sun}, \bibinfo{person}{Ren Wang}, \bibinfo{person}{Yihua Zhang}, \bibinfo{person}{Jinghan Jia}, \bibinfo{person}{Jiancheng Liu}, \bibinfo{person}{Gaowen Liu}, \bibinfo{person}{Yan Yan}, {and} \bibinfo{person}{Sijia Liu}.} \bibinfo{year}{2024}\natexlab{a}.
\newblock \showarticletitle{Forget Vectors at Play: Universal Input Perturbations Driving Machine Unlearning in Image Classification}.
\newblock \bibinfo{journal}{\emph{arXiv preprint arXiv:2412.16780}} (\bibinfo{year}{2024}).
\newblock


\bibitem[Sun et~al\mbox{.}(2024b)]%
        {sun2024r}
\bibfield{author}{\bibinfo{person}{Zhaoyan Sun}, \bibinfo{person}{Xuanhe Zhou}, {and} \bibinfo{person}{Guoliang Li}.} \bibinfo{year}{2024}\natexlab{b}.
\newblock \showarticletitle{R-Bot: An LLM-based Query Rewrite System}.
\newblock \bibinfo{journal}{\emph{arXiv preprint arXiv:2412.01661}} (\bibinfo{year}{2024}).
\newblock


\bibitem[Tang et~al\mbox{.}(2023)]%
        {tang2023does}
\bibfield{author}{\bibinfo{person}{Ruixiang Tang}, \bibinfo{person}{Xiaotian Han}, \bibinfo{person}{Xiaoqian Jiang}, {and} \bibinfo{person}{Xia Hu}.} \bibinfo{year}{2023}\natexlab{}.
\newblock \showarticletitle{Does synthetic data generation of llms help clinical text mining?}
\newblock \bibinfo{journal}{\emph{arXiv preprint arXiv:2303.04360}} (\bibinfo{year}{2023}).
\newblock


\bibitem[Thaker et~al\mbox{.}(2024)]%
        {thaker2024guardrail}
\bibfield{author}{\bibinfo{person}{Pratiksha Thaker}, \bibinfo{person}{Yash Maurya}, {and} \bibinfo{person}{Virginia Smith}.} \bibinfo{year}{2024}\natexlab{}.
\newblock \showarticletitle{Guardrail Baselines for Unlearning in LLMs}.
\newblock \bibinfo{journal}{\emph{arXiv preprint arXiv:2403.03329}} (\bibinfo{year}{2024}).
\newblock


\bibitem[Touvron et~al\mbox{.}(2023)]%
        {touvron2023llama}
\bibfield{author}{\bibinfo{person}{Hugo Touvron}, \bibinfo{person}{Louis Martin}, \bibinfo{person}{Kevin Stone}, \bibinfo{person}{Peter Albert}, \bibinfo{person}{Amjad Almahairi}, \bibinfo{person}{Yasmine Babaei}, \bibinfo{person}{Nikolay Bashlykov}, \bibinfo{person}{Soumya Batra}, \bibinfo{person}{Prajjwal Bhargava}, \bibinfo{person}{Shruti Bhosale}, {et~al\mbox{.}}} \bibinfo{year}{2023}\natexlab{}.
\newblock \showarticletitle{Llama 2: Open foundation and fine-tuned chat models}.
\newblock \bibinfo{journal}{\emph{arXiv preprint arXiv:2307.09288}} (\bibinfo{year}{2023}).
\newblock


\bibitem[Tunstall et~al\mbox{.}(2023)]%
        {tunstall2023zephyr}
\bibfield{author}{\bibinfo{person}{Lewis Tunstall}, \bibinfo{person}{Edward Beeching}, \bibinfo{person}{Nathan Lambert}, \bibinfo{person}{Nazneen Rajani}, \bibinfo{person}{Kashif Rasul}, \bibinfo{person}{Younes Belkada}, \bibinfo{person}{Shengyi Huang}, \bibinfo{person}{Leandro von Werra}, \bibinfo{person}{Clémentine Fourrier}, \bibinfo{person}{Nathan Habib}, \bibinfo{person}{Nathan Sarrazin}, \bibinfo{person}{Omar Sanseviero}, \bibinfo{person}{Alexander~M. Rush}, {and} \bibinfo{person}{Thomas Wolf}.} \bibinfo{year}{2023}\natexlab{}.
\newblock \bibinfo{title}{Zephyr: Direct Distillation of LM Alignment}.
\newblock
\showeprint[arxiv]{2310.16944}~[cs.LG]


\bibitem[Wei et~al\mbox{.}(2024)]%
        {wei2024assessing}
\bibfield{author}{\bibinfo{person}{Boyi Wei}, \bibinfo{person}{Kaixuan Huang}, \bibinfo{person}{Yangsibo Huang}, \bibinfo{person}{Tinghao Xie}, \bibinfo{person}{Xiangyu Qi}, \bibinfo{person}{Mengzhou Xia}, \bibinfo{person}{Prateek Mittal}, \bibinfo{person}{Mengdi Wang}, {and} \bibinfo{person}{Peter Henderson}.} \bibinfo{year}{2024}\natexlab{}.
\newblock \showarticletitle{Assessing the brittleness of safety alignment via pruning and low-rank modifications}.
\newblock \bibinfo{journal}{\emph{arXiv preprint arXiv:2402.05162}} (\bibinfo{year}{2024}).
\newblock


\bibitem[Wei and Zou(2019)]%
        {wei2019eda}
\bibfield{author}{\bibinfo{person}{Jason Wei} {and} \bibinfo{person}{Kai Zou}.} \bibinfo{year}{2019}\natexlab{}.
\newblock \showarticletitle{Eda: Easy data augmentation techniques for boosting performance on text classification tasks}.
\newblock \bibinfo{journal}{\emph{arXiv preprint arXiv:1901.11196}} (\bibinfo{year}{2019}).
\newblock


\bibitem[Wen et~al\mbox{.}(2023)]%
        {wen2023unveiling}
\bibfield{author}{\bibinfo{person}{Jiaxin Wen}, \bibinfo{person}{Pei Ke}, \bibinfo{person}{Hao Sun}, \bibinfo{person}{Zhexin Zhang}, \bibinfo{person}{Chengfei Li}, \bibinfo{person}{Jinfeng Bai}, {and} \bibinfo{person}{Minlie Huang}.} \bibinfo{year}{2023}\natexlab{}.
\newblock \showarticletitle{Unveiling the Implicit Toxicity in Large Language Models}. In \bibinfo{booktitle}{\emph{The 2023 Conference on Empirical Methods in Natural Language Processing}}.
\newblock


\bibitem[Wu et~al\mbox{.}(2023b)]%
        {wu2023depn}
\bibfield{author}{\bibinfo{person}{Xinwei Wu}, \bibinfo{person}{Junzhuo Li}, \bibinfo{person}{Minghui Xu}, \bibinfo{person}{Weilong Dong}, \bibinfo{person}{Shuangzhi Wu}, \bibinfo{person}{Chao Bian}, {and} \bibinfo{person}{Deyi Xiong}.} \bibinfo{year}{2023}\natexlab{b}.
\newblock \showarticletitle{DEPN: Detecting and Editing Privacy Neurons in Pretrained Language Models}.
\newblock \bibinfo{journal}{\emph{arXiv preprint arXiv:2310.20138}} (\bibinfo{year}{2023}).
\newblock


\bibitem[Wu et~al\mbox{.}(2023a)]%
        {wu2023dipmark}
\bibfield{author}{\bibinfo{person}{Yihan Wu}, \bibinfo{person}{Zhengmian Hu}, \bibinfo{person}{Hongyang Zhang}, {and} \bibinfo{person}{Heng Huang}.} \bibinfo{year}{2023}\natexlab{a}.
\newblock \showarticletitle{Dipmark: A stealthy, efficient and resilient watermark for large language models}.
\newblock  (\bibinfo{year}{2023}).
\newblock


\bibitem[Yao et~al\mbox{.}(2024a)]%
        {yao2024machine}
\bibfield{author}{\bibinfo{person}{Jin Yao}, \bibinfo{person}{Eli Chien}, \bibinfo{person}{Minxin Du}, \bibinfo{person}{Xinyao Niu}, \bibinfo{person}{Tianhao Wang}, \bibinfo{person}{Zezhou Cheng}, {and} \bibinfo{person}{Xiang Yue}.} \bibinfo{year}{2024}\natexlab{a}.
\newblock \showarticletitle{Machine Unlearning of Pre-trained Large Language Models}.
\newblock \bibinfo{journal}{\emph{arXiv preprint arXiv:2402.15159}} (\bibinfo{year}{2024}).
\newblock


\bibitem[Yao et~al\mbox{.}(2024b)]%
        {yao2023large}
\bibfield{author}{\bibinfo{person}{Yuanshun Yao}, \bibinfo{person}{Xiaojun Xu}, {and} \bibinfo{person}{Yang Liu}.} \bibinfo{year}{2024}\natexlab{b}.
\newblock \showarticletitle{Large Language Model Unlearning}. In \bibinfo{booktitle}{\emph{The Thirty-eighth Annual Conference on Neural Information Processing Systems}}.
\newblock


\bibitem[Yu et~al\mbox{.}(2023a)]%
        {yu2023unlearning}
\bibfield{author}{\bibinfo{person}{Charles Yu}, \bibinfo{person}{Sullam Jeoung}, \bibinfo{person}{Anish Kasi}, \bibinfo{person}{Pengfei Yu}, {and} \bibinfo{person}{Heng Ji}.} \bibinfo{year}{2023}\natexlab{a}.
\newblock \showarticletitle{Unlearning bias in language models by partitioning gradients}. In \bibinfo{booktitle}{\emph{Findings of the Association for Computational Linguistics: ACL 2023}}. \bibinfo{pages}{6032--6048}.
\newblock


\bibitem[Yu et~al\mbox{.}(2023b)]%
        {yu2023large}
\bibfield{author}{\bibinfo{person}{Yue Yu}, \bibinfo{person}{Yuchen Zhuang}, \bibinfo{person}{Jieyu Zhang}, \bibinfo{person}{Yu Meng}, \bibinfo{person}{Alexander~J Ratner}, \bibinfo{person}{Ranjay Krishna}, \bibinfo{person}{Jiaming Shen}, {and} \bibinfo{person}{Chao Zhang}.} \bibinfo{year}{2023}\natexlab{b}.
\newblock \showarticletitle{Large language model as attributed training data generator: A tale of diversity and bias}.
\newblock \bibinfo{journal}{\emph{Advances in Neural Information Processing Systems}}  \bibinfo{volume}{36} (\bibinfo{year}{2023}), \bibinfo{pages}{55734--55784}.
\newblock


\bibitem[Zhang et~al\mbox{.}(2024b)]%
        {zhang2024negative}
\bibfield{author}{\bibinfo{person}{Ruiqi Zhang}, \bibinfo{person}{Licong Lin}, \bibinfo{person}{Yu Bai}, {and} \bibinfo{person}{Song Mei}.} \bibinfo{year}{2024}\natexlab{b}.
\newblock \showarticletitle{Negative Preference Optimization: From Catastrophic Collapse to Effective Unlearning}. In \bibinfo{booktitle}{\emph{First Conference on Language Modeling}}.
\newblock


\bibitem[Zhang et~al\mbox{.}(2024a)]%
        {zhang2024generate}
\bibfield{author}{\bibinfo{person}{Yimeng Zhang}, \bibinfo{person}{Jinghan Jia}, \bibinfo{person}{Xin Chen}, \bibinfo{person}{Aochuan Chen}, \bibinfo{person}{Yihua Zhang}, \bibinfo{person}{Jiancheng Liu}, \bibinfo{person}{Ke Ding}, {and} \bibinfo{person}{Sijia Liu}.} \bibinfo{year}{2024}\natexlab{a}.
\newblock \showarticletitle{To generate or not? safety-driven unlearned diffusion models are still easy to generate unsafe images... for now}. In \bibinfo{booktitle}{\emph{European Conference on Computer Vision}}. Springer, \bibinfo{pages}{385--403}.
\newblock


\bibitem[Zhao et~al\mbox{.}(2023)]%
        {zhao2023provable}
\bibfield{author}{\bibinfo{person}{Xuandong Zhao}, \bibinfo{person}{Prabhanjan Ananth}, \bibinfo{person}{Lei Li}, {and} \bibinfo{person}{Yu-Xiang Wang}.} \bibinfo{year}{2023}\natexlab{}.
\newblock \showarticletitle{Provable robust watermarking for ai-generated text}.
\newblock \bibinfo{journal}{\emph{arXiv preprint arXiv:2306.17439}} (\bibinfo{year}{2023}).
\newblock


\bibitem[Zhuang et~al\mbox{.}(2024)]%
        {zhuang2024uoe}
\bibfield{author}{\bibinfo{person}{Haomin Zhuang}, \bibinfo{person}{Yihua Zhang}, \bibinfo{person}{Kehan Guo}, \bibinfo{person}{Jinghan Jia}, \bibinfo{person}{Gaowen Liu}, \bibinfo{person}{Sijia Liu}, {and} \bibinfo{person}{Xiangliang Zhang}.} \bibinfo{year}{2024}\natexlab{}.
\newblock \showarticletitle{UOE: Unlearning One Expert Is Enough For Mixture-of-experts LLMS}.
\newblock \bibinfo{journal}{\emph{arXiv preprint arXiv:2411.18797}} (\bibinfo{year}{2024}).
\newblock


\bibitem[Zubiaga(2024)]%
        {zubiaga2024natural}
\bibfield{author}{\bibinfo{person}{Arkaitz Zubiaga}.} \bibinfo{year}{2024}\natexlab{}.
\newblock \bibinfo{title}{Natural language processing in the era of large language models}.
\newblock \bibinfo{numpages}{1350306}~pages.
\newblock


\end{thebibliography}

\clearpage 
\newpage
\clearpage
\onecolumn
\section*{\Large{Appendix}}
\setcounter{section}{0}
\setcounter{figure}{0}
\setcounter{table}{0}
\makeatletter 
\renewcommand{\thesection}{\Alph{section}}
\renewcommand{\theHsection}{\Alph{section}}
\renewcommand{\thefigure}{A\arabic{figure}}
\renewcommand{\theHfigure}{A\arabic{figure}}
\renewcommand{\thetable}{A\arabic{table}}
\renewcommand{\theHtable}{A\arabic{table}}
\makeatother

\renewcommand{\thetable}{A\arabic{table}}
\setcounter{mylemma}{0}
\renewcommand{\themylemma}{A\arabic{mylemma}}
\setcounter{equation}{0}
\renewcommand{\theequation}{A\arabic{equation}}

\section{Experiment Setup and Implementation Details}
\label{appendix: setup}

\subsection{Unlearning Configurations}

\subsubsection{WMDP benchmark} We use the forget set provided in the WMDP~\cite{li2024wmdp} benchmark, which contains a large collection of biology-related articles. For the retain set, we select WikiText~\cite{merity2016pointer}, whose content is presumed unrelated to the forget set. Our baseline model is Zephyr-7B-beta, as specified in the WMDP benchmark. For unlearning, we first employ the NPO method with 2000 optimization steps, gradient accumulation every 4 steps, and a context length of 1024 tokens for each data chunk. The learning rate is chosen via a grid search in $[10^{-6}, 10^{-5}]$, while the parameter~$\gamma$ appearing before the retain loss is selected from $[1, 2.5]$. We choose the final unlearned model as the one that preserves performance closest to the original Zephyr-7B-beta. We also employ the RMU method, using a batch size of 4 and sampling 800 total data instances, each with 512 tokens per data chunk. The learning rate is tuned within $[10^{-5}, 10^{-3}]$, and the parameter~$\alpha$ appearing before the retain loss is searched in $[1, 10]$.

\subsubsection{MUSE benchmark} For MUSE~\citep{shi2024muse}, we adopt {ICLM 7B} fine-tuned on Harry Potter books as the base model, is trained for 1 epochs with a learning rate of $10^{-5}$, and we set $\beta = 0.1$. Following prior work, we perform grid search for the regularization coefficient $\lambda$ before $\ell_r$ within the range $[0.25, 1.0]$. The same configuration is applied across all forget data types.

\subsection{Error Set Overlap}
\label{appendix:Error_Set_overlap}

To quantify the consistency of forgetting behavior under different forget data perturbations, we define the \textbf{Error Set Overlap Ratio} as a measure of semantic alignment between unlearned models.

Let $\mathcal{E}_\mathrm{orig}$ denote the \textit{error set} of the model unlearned with the original forget data $\mathcal{D}_\mathrm{f}$, and $\mathcal{E}_\mathrm{pert}$ the error set of the model unlearned with a perturbed variant $\mathcal{D}_\mathrm{f}'$. Each error set is defined as the set of questions in the WMDP evaluation QA set that are \textbf{answered incorrectly} by the corresponding unlearned model.
We then compute the \textit{Error Set Overlap Ratio} between the two models as the Jaccard similarity between their error sets:
\begin{align}
\mathrm{Error\ Set\ Overlap\ Ratio}(\mathcal{E}_\mathrm{orig}, \mathcal{E}_\mathrm{pert}) = 
\frac{|\mathcal{E}_\mathrm{orig} \cap \mathcal{E}_\mathrm{pert}|}
{|\mathcal{E}_\mathrm{orig} \cup \mathcal{E}_\mathrm{pert}|}.
\end{align}
This ratio captures the extent to which the two models forget the same underlying knowledge. A higher overlap ratio indicates that the perturbed forget data results in forgetting effects similar to those produced by the original data.

\subsection{Salient Tokens Extraction}
\label{appendix:keywords}

To complement the analysis of unlearning consistency under perturbed forget data, we define the \textbf{Salient Tokens Overlap Ratio} as a metric to quantify semantic alignment at the keyword level.

We begin by extracting concept-relevant \textit{salient tokens} from each forget sample using a prompt-based LLM-as-a-judge framework (see prompt in Appendix~\ref{prompt:meta}). The extraction is conducted using the \texttt{GPT-o3-mini} model, which takes the forget sample as input and returns a list of key concepts or entities central to the sentence meaning.

Let $K_\mathrm{orig}$ denote the set of  salient tokens extracted from the original forget dataset $\mathcal{D}_\mathrm{f}$, and $K_\mathrm{pert}$ the corresponding salient tokens from the perturbed dataset $\mathcal{D}_\mathrm{f}'$. We then define the \textit{Salient Tokens Overlap Ratio} as the Jaccard similarity between these two keyword sets:
\begin{align}
\mathrm{Salient \ Tokens \ Overlap \ Ratio}(K_\mathrm{orig}, K_\mathrm{pert}) = 
\frac{|K_\mathrm{orig} \cap K_\mathrm{pert}|}
{|K_\mathrm{orig} \cup K_\mathrm{pert}|}.
\end{align}
This metric captures the extent to which the semantic core of the original data is preserved in its perturbed variant. A high salient tokens overlap ratio indicates that the perturbation retains the key semantic signals necessary for effective unlearning.

\newtcolorbox{prompt}[2][]{colback=gray!5!white,colframe=gray!75!black,title=#2,#1}

\begin{prompt}{Prompt to Extract Salient Tokens using LLM-as-a-Judge}\label{prompt:meta}
\textbf{System}:
\\
\\
You are an expert in chemistry, biology and diseases. I will provide you with a text, and your task is to carefully analyze each word and determine if it is related to chemistry or biology and disease. For every word that is related to chemistry, biology and disease, replace it with an asterisk (*) in the original text while keeping all other words unchanged. Think step by step and ensure accuracy in identifying only chemistry, biology and disease-related terms. Pay close attention to scientific terminology. Return the result in the following format:\\
\{\{ \\
"replaced\_text": "$<$text after replacing$>$",\\
"related\_words": ["word1", "word2", "word3", ...] \\
\}\} \\
Do \textbf{not} write any code. Use your linguistic and scientific knowledge to analyze the text.
\\
\\
\textbf{User}: 
\\
\\
{\color{blue}\{WMDP-Bio forget set\}}
\\
\\
\textbf{Assistant}:
\\
\\
{\color{blue}\{response\}}

\end{prompt}


\appendix


\end{document}